%% file: medima-template.tex
\documentclass[times,twocolumn,final]{elsarticle}

\usepackage{medima}
\usepackage{framed,multirow}

\usepackage{amssymb}
\usepackage{latexsym}

\usepackage{url}
\usepackage{xcolor}
\usepackage{array}
\usepackage{arydshln}
\usepackage{subcaption}
\usepackage{hyperref}
\usepackage{makecell}
\usepackage{booktabs}
\usepackage{multirow}
\usepackage{caption}
\usepackage{bbding}
\usepackage{graphicx}
\usepackage{hyperref}
\usepackage{array}
\usepackage{arydshln}
\usepackage{subcaption}
\usepackage{hyperref}
\usepackage{makecell}
\usepackage{booktabs}
\usepackage{multirow}
\usepackage{amsmath}
\usepackage{caption}
\usepackage{graphicx}
\usepackage{bm}
\definecolor{newcolor}{rgb}{.8,.349,.1}

\journal{Medical Image Analysis}

\begin{document}

\verso{Given-name Surname \textit{et~al.}}

\begin{frontmatter}

\title{MedLSAM: Localize and Segment Anything Model for 3D CT Images}

\author[1,2]{Wenhui \snm{Lei}}
\author[3,4]{Wei \snm{Xu}}
\author[2,4]{Kang \snm{Li}}
\author[1,2]{Xiaofan \snm{Zhang}\corref{cor1}}
\author[1,2]{Shaoting \snm{Zhang}}
\cortext[cor1]{Corresponding author: xiaofan.zhang@sjtu.edu.cn}
\address[1]{School of Electronic Information and Electrical Engineering, Shanghai Jiao Tong University, Shanghai, China}
\address[2]{Shanghai AI Lab, Shanghai, China}
\address[3]{School of Biomedical Engineering, Division of Life Sciences and Medicine, University of Science and Technology of China, Hefei, China}
\address[4]{West China Biomedical Big Data Center, West China Hospital, Sichuan University, Chengdu, China}

\begin{abstract}
\textcolor{black}{Recent advancements in foundation models have shown significant potential in medical image analysis. However, there is still a gap in models specifically designed for medical image localization. To address this, we introduce MedLAM, a 3D medical foundation localization model that accurately identifies any anatomical part within the body using only a few template scans. MedLAM employs two self-supervision tasks: unified anatomical mapping (UAM) and multi-scale similarity (MSS) across a comprehensive dataset of 14,012 CT scans. Furthermore, we developed MedLSAM by integrating MedLAM with the Segment Anything Model (SAM). This innovative framework requires extreme point annotations across three directions on several templates to enable MedLAM to locate the target anatomical structure in the image, with SAM performing the segmentation. It significantly reduces the amount of manual annotation required by SAM in 3D medical imaging scenarios. We conducted extensive experiments on two 3D datasets covering 38 distinct organs. Our findings are twofold: 1) MedLAM can directly localize anatomical structures using just a few template scans, achieving performance comparable to fully supervised models; 2) MedLSAM closely matches the performance of SAM and its specialized medical adaptations with manual prompts, while minimizing the need for extensive point annotations across the entire dataset. Moreover, MedLAM has the potential to be seamlessly integrated with future 3D SAM models, paving the way for enhanced segmentation performance.} Our code is public at \href{https://github.com/openmedlab/MedLSAM}{https://github.com/openmedlab/MedLSAM}.
\end{abstract}


\end{frontmatter}


\section{Introduction}
\label{sec1}

\par Recently, there has been an increasing interest in the field of computer vision to develop large-scale \textit{foundation models} that can concurrently address multiple visual tasks, such as image classification, object detection, and image segmentation. For instance, CLIP \citep{radford2021learning}, aligning a vast number of images and text pairs collected from the web, could recognize new visual categories using text \textit{prompts}. Similarly, GLIP \citep{li2022grounded, zhang2022glipv2}, which integrates object detection with phrase grounding in pre-training using web-sourced image-text pairs, boasts strong zero-shot and few-shot transferability in object-level detection tasks. On the segmentation front, the \textit{Segment Anything Model (SAM)} \citep{kirillov2023segment} has recently demonstrated remarkable capabilities in a broad range of segmentation tasks with appropriate prompts, e.g. points, bounding box (bbox) and text.

\par As foundation models increasingly demonstrate their prowess in general computer vision tasks, the medical imaging domain, characterized by limited image availability and high annotation costs, is taking keen notice of their potential. Such challenges in medical imaging underscore the pressing need for foundation models, drawing greater attention from researchers \citep{zhang2023challenges}. Some studies have endeavored to design self-supervised learning tasks tailored to dataset characteristics, pre-training on a vast amount of unlabeled data, and then fine-tuning on specific downstream tasks to achieve commendable results \citep{wang2023mis, zhou2023foundation, vorontsov2023virchow}. In contrast, other works have focused on pre-training with large annotated datasets, either to fine-tune on new tasks \citep{huang2023stu} or to enable direct segmentation without the necessity of further training or fine-tuning \citep{butoi2023universeg}. Overall, these models offer myriad advantages: they simplify the development process, reduce dependence on voluminously labeled datasets, and bolster patient data privacy.

\par \textcolor{black}{As a crucial component of medical image analysis, medical image localization encompasses tasks such as keypoint detection\citep{chen2021fast, wan2023multi}, organ localization\citep{xu2019efficient,navarro2020deep, hussain2021cascaded, navarro2022unified}, and disease detection\citep{yan2018deeplesion, zhang2018automatic}. However, to the best of our knowledge, foundation models specifically designed for medical image localization tasks remain very limited.  In the domain of 2D medical image localization, MIU-VL \citep{qin2022medical} stands out as a pioneering effort.  This approach ingeniously marries pre-trained vision-language models with medical text prompts to facilitate object detection in medical imaging. However, the application of these pre-trained models to specialized medical imaging domains, particularly 3D modalities like CT scans, remains challenging. This limitation is attributable to a significant domain gap, as these models were originally trained on 2D natural images.}

\par \textcolor{black}{At the same time, in the field of medical image segmentation,  models such as SAM and its medical adaptations \citep{MSA, cheng2023sam, he2023sam, huang2023segment, he2023accuracy, zhang2023customized, mazurowski2023segment, zhang2023segment} have demonstrated significant potential, even achieving performance comparable to fully supervised segmentation models in certain tasks \citep{MedSAM}.} Despite these advancements, these models still require manual prompts, such as labeled points or bounding boxes. However, a single 3D scan often contains dozens or even hundreds of slices. When each slice comprises multiple target structures, the annotation \textcolor{black}{demands} for the entire 3D scan grow exponentially \citep{wang2018deepigeos, lei2019deepigeos}. To eliminate this vast annotation demand and facilitate the use of SAM and its medical variants on 3D data, an effective approach would be to automatically generate suitable prompts for the data awaiting annotation. While this could be achieved by training a detector specifically for the categories to be annotated, it introduces the additional burden of annotating data for detector training, and the detectable categories remain fixed \citep{baumgartner2021nndetection}. A more optimal solution would be to develop a foundation localization model for 3D medical images that only requires minimal user input to specify the category of interest, allowing for direct localization of the desired target in any dataset without further training. \textcolor{black}{This challenge underscores the urgent need for a 3D medical localization foundation model.}

\begin{figure}[t!]
    \centering
    \includegraphics[width=1\linewidth]{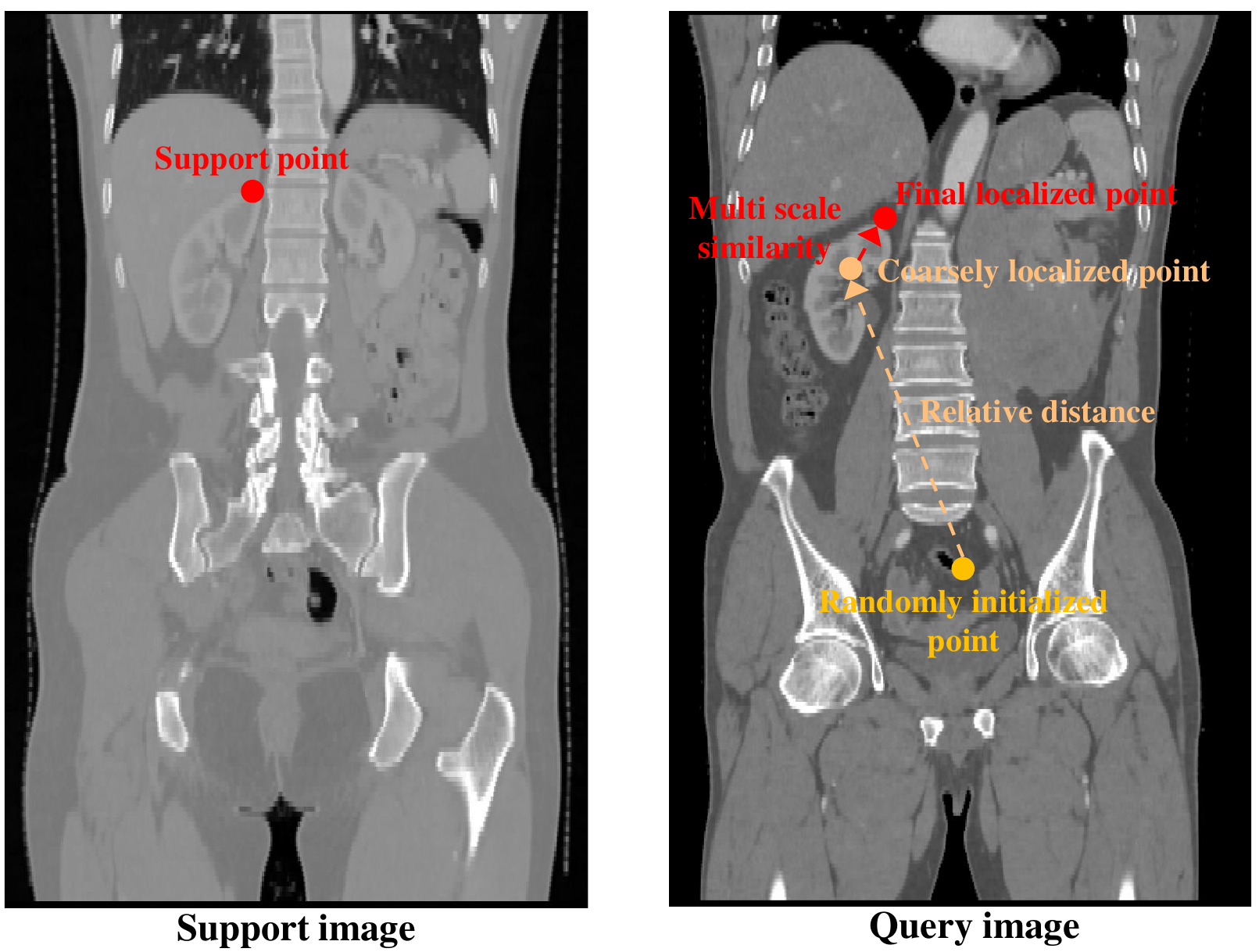}
    \caption{Structure of the inference stage in the Medical Localization Anything Model (MedLAM). The process involves moving an agent from a randomly initialized position toward the target landmark, guided by 3D relative distance and multi-scale feature vectors.}
    \label{fig:inference}
\end{figure} 

\par \textcolor{black}{To address the lack of foundation models for 3D medical image localization, we introduce the \textbf{Localize Anything Model for 3D Medical Images (MedLAM)}. MedLAM is a specialized 3D localization foundation model designed to accurately locate any anatomical structure using minimal prompts. Additionally, we integrated MedLAM with the SAM model to create \textbf{MedLSAM}, a framework that significantly reduces the need for extensive manual interaction in SAM while demonstrating the potential to achieve competitive performance. } 
\par MedLSAM employs a two-stage methodology. The first stage involves MedLAM to automatically identify the positions of target structures within volumetric medical images. In the subsequent stage, the SAM model uses the bboxes from the first stage to achieve precise image segmentation. The result is a fully autonomous pipeline that \textcolor{black}{minimizes} the need for manual intervention.

The localization foundation model, MedLAM, is an extension of our conference version \citep{lei2021contrastive} and is premised on the observation that the spatial distribution of organs maintains strong similarities across different individuals. And we assume that there exists a standard anatomical coordinate system, in which the same anatomy part in different people shares similar coordinates. Therefore, we can localize the targeted anatomy structure with a similar coordinate in the unannotated scan. In our previous version, the model was trained separately for different anatomical structures, each using only a few dozen scan images. In contrast, this current study significantly expands the dataset to include 14,012 CT scans from 16 different datasets. This allows us to train a unified, comprehensive model capable of localizing any structure across the entire body. The training process involves a projection network that predicts the 3D physical offsets between any two patches within the same image, thereby mapping every part of the scans onto a shared 3D latent coordinate system.

For the segmentation stage, we employ the original SAM and the well-established MedSAM \citep{MedSAM} as the foundation for our segmentation process. MedSAM, previously fine-tuned on a comprehensive collection of medical image datasets, has exhibited considerable performance in 2D and 3D medical image segmentation tasks. The use of such a robust model bolsters the reliability and effectiveness of our proposed pipeline.

Our contributions can be summarized as follows:
\par \textcolor{black}{-  We introduce MedLAM, the first 3D medical image localization foundation model. This model, an extension of our prior conference work \citep{lei2021contrastive}, incorporates pixel-level self-supervised tasks and is trained on a comprehensive dataset of 14,012 CT scans covering the entire body. As a result, it's capable of directly localizing any anatomical structure within the human body and even comparable to the fully-supervised localization models in most cases;} 
\par - We introduce MedLSAM, the first completely automated medical adaptation of the SAM model. It not only \textcolor{black}{minimizes} the need for manual intervention but also propels the development of fully automated foundation segmentation models in medical image analysis. To the best of our knowledge, this is the first work to achieve complete automation in medical image segmentation by integrating the SAM model;
\par - We validate the effectiveness of \textcolor{black}{MedLAM and} MedLSAM on two 3D datasets that span 38 organs. As showcased in \textcolor{black}{Table \ref{tab:2} and Table \ref{tab:3}}, our localization foundation model, MedLAM, exhibits localization accuracy that not only significantly outperforms existing medical localization foundation models but also rivals the performance of fully-supervised models. Concurrently, as demonstrated in Table \ref{tab:4}, MedLSAM aligns with the performance of SAM and its medical adaptations, substantially \textcolor{black}{minimizing} the reliance on manual annotations. To further the research in foundation models for medical imaging, we have made all our models and codes publicly available.

\section{Related Work}
\subsection{Medical Visual Foundation Models}

\par Over recent years, large-scale pre-trained deep learning vision models have marked remarkable milestones with models like ResNet\citep{he2016deep}, ConvNeXt\citep{liu2022convnet}, Vision Transformers (ViT)\citep{dosovitskiy2020image}, and DINO\citep{caron2021emerging}. Owing to their profound representational power and exceptional generalization capabilities, these models have consistently excelled in various downstream applications.
\par In the realm of medical image analysis, the challenges are compounded by concerns of data privacy and the often tedious nature of annotation tasks, making the procurement of large-scale annotated data both challenging and costly. This elevates the significance of visual foundation models in the field. For instance, REMEDIS\citep{azizi2022robust} harnesses the potential of self-supervised learning on a diverse set of medical images spanning multiple modalities. Impressively, it rivals the performance of heavily supervised baselines in downstream disease classification tasks, requiring only between 1\% to 33\% of the typical retraining data. Meanwhile, SAIS\citep{azizi2022robust}, capitalizing on the foundation model DINO\citep{caron2021emerging} pre-trained on standard images, showcases impressive generalization and prowess in specialized tasks, such as surgical gesture recognition and skill evaluation.
\par Turning the attention to segmentation, a key aspect within medical imaging, there has been a surge of interest in the application of SAM for medical image segmentation \citep{cheng2023sam, he2023sam, MSA, MedSAM, huang2023segment, he2023accuracy, zhang2023customized, mazurowski2023segment, zhang2023segment}. \textcolor{black}{Both \cite{huang2023segment} and \cite{cheng2023sam} have collected extensive medical image datasets and conducted comprehensive testing on SAM. They discovered that, while SAM exhibits remarkable performance on certain objects and modalities, it is less effective or even fails in other scenarios. This inconsistency arises from SAM's primary training on natural images. To enhance SAM's generalization in medical imaging, studies like \cite{MSA, MedSAM} have fine-tuned the original SAM using a large annotated medical image dataset, thereby validating its improved performance over the base SAM model in test datasets. } 

However, a significant hurdle for SAM's utility in medical image segmentation is the necessity for manual annotations. This involves annotating specific points or bboxes that demarcate the segmentation region, a process that is both time-consuming and expensive. Intriguingly, comprehensive studies by \cite{huang2023segment} and \cite{cheng2023sam} have delved into the optimal prompts for SAM's application in medical imaging. They primarily investigated prompt types including bboxes, a combination of bboxes with foreground and background points, and the exclusive use of foreground and background points. Notably, in the majority of tasks, bboxes alone showcased superior performance.

In light of these findings, one of the central focuses of this paper is the automation of generating bbox prompts for SAM. Furthermore, we aim to advance the full automation of SAM's application in the medical domain, reducing the reliance on manual annotations and propelling its utility.

\begin{figure*}[t!]
\centering
\includegraphics[width=0.85\linewidth]{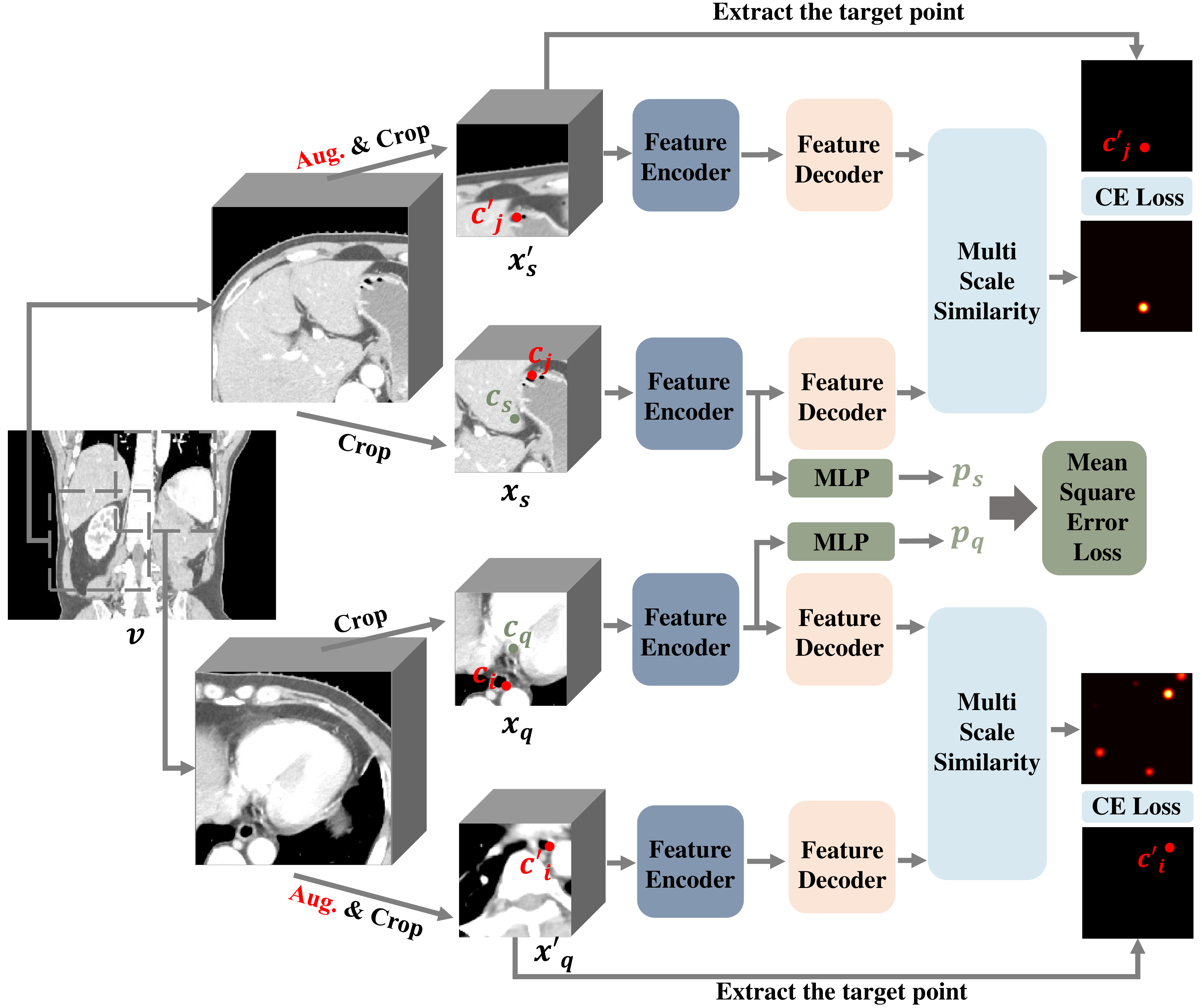}
\caption{The learning process of MedLAM. \textcolor{black}{We first randomly extract two large image patches from the scan. Then we randomly crop a small image patch from each of the two large image patches and their augmented versions to produce two pairs of patches, namely, the original patch pair $\bm{x}_q, \bm{x}_s$ and the augmented patch pair $\bm{x_q'}, \bm{x_s'}$. All these patches are passed through the MedLAM and the training objectives contain twofold: 
1) \textbf{Unified Anatomical Mapping (UAM)}: By predicting the relative distance between the original image patches $\bm{x}_q$ and $\bm{x}_s$, MedLAM project images from different individuals onto a shared anatomical coordinate space.
2) \textbf{Multi-Scale Similarity (MSS)}: Ensure that the similarity between features corresponding to the same region is maximized in both the original image patches $\bm{x}_q, \bm{x}_s$ and their augmented counterparts $\bm{x_q'}, \bm{x_s'}$.
}}
\label{fig:training}
\end{figure*}

\subsection{Medical Localization Models}
Traditionally, localization tasks, such as object detection, have relied on extensive annotations for a limited set of categories to enable models to produce corresponding bboxes \citep{redmon2016you, he2017mask, dai2016r, ren2015faster}. This challenge has also been extensively addressed in the realm of medical imaging, with a focus on areas such as organ localization \citep{xu2019efficient,navarro2020deep, hussain2021cascaded, navarro2022unified}, keypoint detection\citep{chen2021fast, wan2023multi}, and lesion detection\citep{yan2018deeplesion, zhang2018automatic}. Both organ localization and keypoint detection often serve as preliminary steps for many medical image analysis tasks, including image registration\citep{sun2013validation}, organ segmentation\citep{liu2023clip}, and disease diagnosis\citep{bilic2023liver}. While previous fully-supervised models have achieved commendable results in predefined tasks, their primary limitation is their difficulty in generalizing to new categories. The high cost and challenges associated with data collection and annotation further exacerbate this issue, especially pronounced in the field of medical imaging.
\par To reduce annotation costs, our previous work, the Relative Position Regression-Net (RPR-Net) \citep{lei2021contrastive}, along with another study, Cascade Comparing to Detect (CC2D) \citep{yao2021one}, pioneered the localization of regions of interest or landmarks using annotations from just a single instance. Specifically, CC2D learns to project pixels associated with the same anatomical structures into embeddings with high cosine similarities by addressing a self-supervised "patch matching" proxy task in a coarse-to-fine manner. However, both RPR-Net and CC2D were trained on limited datasets as independent models, without evolving into foundation models for localization. 

\section{Method}

\begin{figure*}[t!]
    \centering
    \includegraphics[width=1\linewidth]{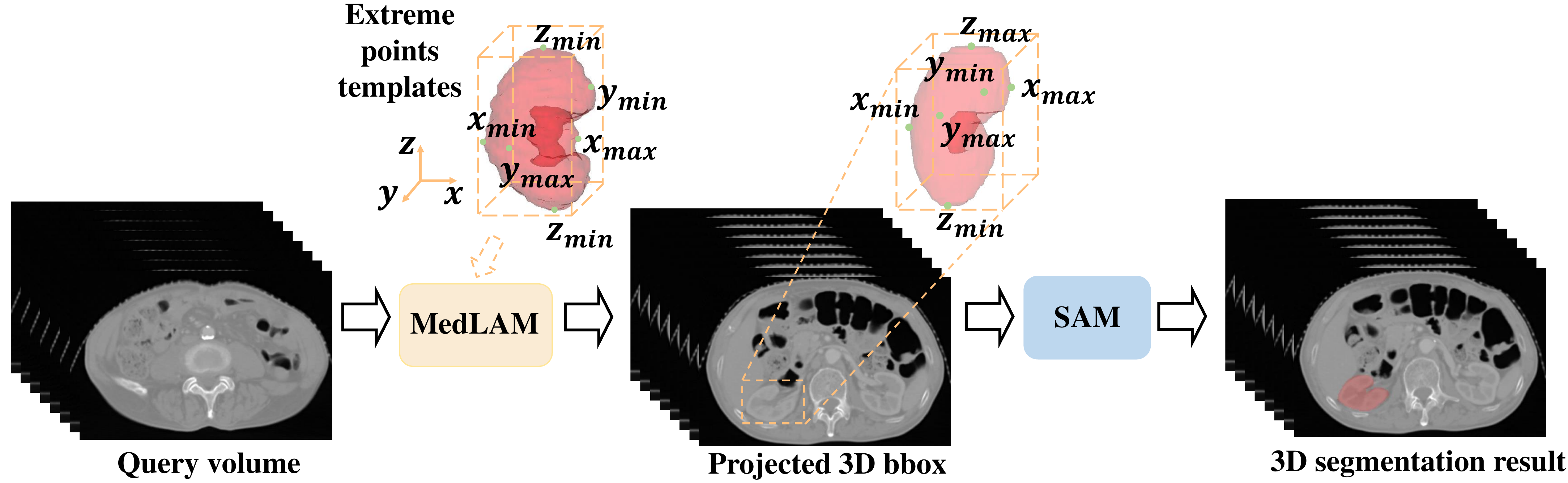}
    \caption{The overall segmentation pipeline of MedLSAM operates as follows. Given a dataset of any size, MedLSAM first applies \textcolor{black}{MedLAM} to identify the six extreme points (in the z, x, and y directions) of any anatomical structure of interest. This process results in the generation of a 3D bbox encompassing the targeted organ or structure. Subsequently, for each slice within this 3D bbox, a 2D bbox is derived, representing the projection of the 3D bbox onto that specific slice. These 2D bboxes are then utilized by the Segment Anything Model (SAM) to carry out precise segmentation of the target anatomy, thereby automating the entire segmentation process.}
    \label{fig:MedLSAM}
\end{figure*}

As illustrated in Fig. \ref{fig:MedLSAM}, MedLSAM comprises two core components: the automated localization algorithm MedLAM and the automated segmentation algorithm SAM. In the subsequent sections, we first delve into the architecture and training of MedLAM. Following that, we detail how MedLAM synergizes with SAM to realize the complete autonomous segmentation framework.

\begin{figure*}[ht!]
    \centering
    \includegraphics[width=0.8\linewidth]{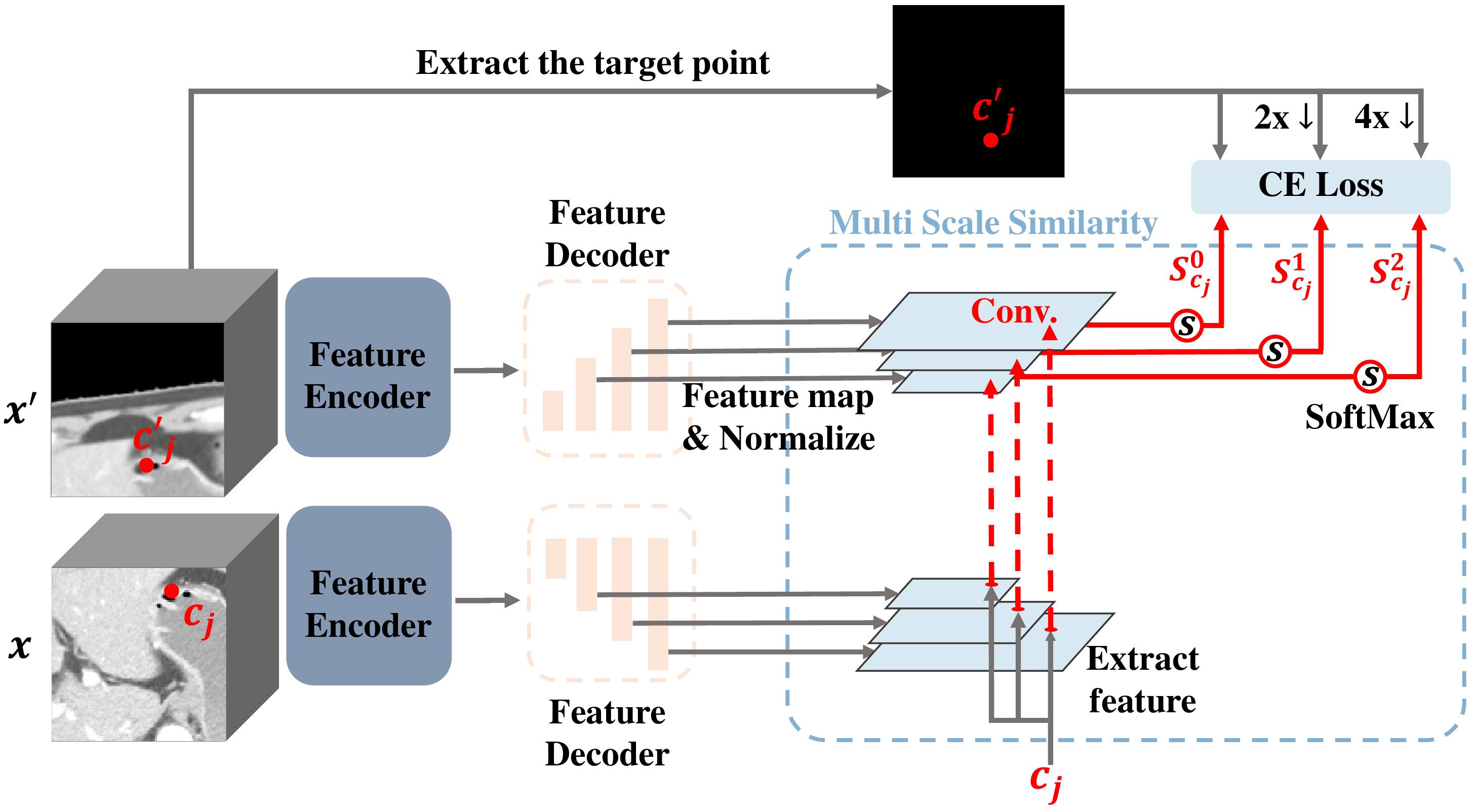}
    \caption{Deatils of the Multi Scale Similarity (MSS). \textcolor{black}{The origin small patch $\bm{x}$ and augmented patch $\bm{x'}$ are all passed through the feature encoder and feature decoder to obtain the normalized multi-scale feature maps. Then we extract the multi-scale feature vectors of point $\bm{c_j}$ to compute the similarity with $\bm{x'}$ corresponding feature maps, applying the softmax operation across all values in each similarity map to achieve the final probability maps set \{$\bm{S_{c_j}^0},\bm{S_{c_j}^1},\bm{S_{c_j}^2}$\}. Finally, the probability maps are restricted with the Cross-Entropy loss.} }
    \label{fig:mss}
\end{figure*}

\subsection{Training of MedLAM}
\label{sec:2.2}

Our MedLAM model, depicted in Fig. \ref{fig:training}, consists of three primary modules: a feature encoder, a multilayer perceptron (MLP), and a feature decoder. Both the feature encoder and decoder are composed of four convolutional blocks. The encoder blocks each have two convolutional layers followed by a downsampling layer, while the decoder blocks contain two convolutional layers succeeded by an upsampling layer. The MLP is structured with three fully connected layers.

\par We start by selecting a volumetric image $\bm{v}$ from the unannotated training set. We then extract two large image patches from $\bm{v}$, which serve as the source for further extractions. These large patches undergo a variety of transformations to produce two pairs of patches, namely, the original patch pair $\bm{x}_q, \bm{x}_s \in R^{D \times W \times H}$ and the augmented patch pair $\bm{x_q'}, \bm{x_s'}\in R^{D \times W \times H}$. The augmentation techniques employed include rotation, elastic transformations, and the addition of random noise. Our training objectives are twofold:
1) \textbf{Unified Anatomical Mapping (UAM)}: To project scan images from different individuals onto a shared anatomical coordinate space by predicting the relative distance between the original image patches $\bm{x}_q$ and $\bm{x}_s$.
2) \textbf{Multi-Scale Similarity (MSS)}: Ensure that the similarity between features corresponding to the same region is maximized in both the original image patches $\bm{x}_q, \bm{x}_s$ and their augmented counterparts $\bm{x_q'}, \bm{x_s'}$.

\subsubsection{Unified Anatomical Mapping (UAM)}
\label{sec:3.1.1}
In this step, we introduce the Unified Anatomical Mapping (UAM) framework, building upon the foundation of our earlier methodology \citep{lei2021contrastive}. The central principle remains consistent: mapping 3D scan images from diverse individuals onto a unified implicit 3D anatomical coordinate system. By doing so, identical anatomical structures from various individuals can be associated with the same coordinate, facilitating an initial, coarse localization of the point within the unannotated scan corresponding to our point of interest.

In contrast to our prior endeavors, UAM embodies a broader vision of creating a model that captures the underlying anatomical structures across a wide range of human body scans. While our previous methods primarily focused on training specialized models for individual localized scan datasets, the UAM framework takes advantage of an extensive collection of CT scans, covering the entire human spectrum. This approach aims to provide a more holistic model capable of operating seamlessly across different organs and body regions, increasing the versatility and applicability of our method but also ensuring higher accuracy and reliability in diverse medical contexts.

The RDR model aims to predict the 3D offset between the query patch $\bm{x}_q$ and the support patch $\bm{x}_s$. Considering $\bm{e} \in R^3$ as the pixel spacing of $\bm{v}$, and $\bm{c}_q, \bm{c}_s \in R^3$ as the centroid coordinates of $\bm{x}_q$ and $\bm{x}_s$ in $\bm{v}$ respectively, the ground truth offset $\bm{d}'_{qs}$ from $\bm{x}_q$ to $\bm{x}_s$ in the physical space can be calculated as:

\begin{equation}
\bm{d}'_{qs}= (\bm{c}_s-\bm{c}_q) \cdot \bm{e} .
\end{equation}

Both $\bm{x}_s$ and $\bm{x}_q$ undergo processing via the feature encoder to distill high-level features. Subsequently, MLP maps these features to their corresponding 3D latent vectors, $\bm{p}_s$ and $\bm{p}_q \in R^3$. This process leads to the predicted offset $\bm{d}_{qs} \in R^3$ from the query patch $\bm{x}_q$ to the support patch $\bm{x}_s$ being computed as:

\begin{equation}
 \bm{d}_{qs}= r \cdot tanh(\bm{p}_{s}-\bm{p}_{q}) .
\end{equation}

The utilization of the hyperbolic tangent function $tanh$ in conjunction with the hyper-parameter $r$ is intended to dictate the upper and lower bound of $\bm{d}_{qs}$, thereby covering the largest feasible offset. Lastly, to measure the difference between $\bm{d}_{qs}$ and $\bm{d}'_{qs}$, we employ the Mean Square Error (MSE) loss function:

\begin{equation}
L_{MSE} = ||\bm{d}_{qs}- \bm{d}'_{qs}||^2 .
\end{equation}

Owing to the constraints set by the hyper-parameter $r$ and the inherent anatomical similarities across different individuals, the model effectively projects the same anatomical regions from various individuals onto similar implicit coordinates after training. This projection mechanism aids in achieving a consistent and unified representation across diverse datasets, thereby enhancing the adaptability and precision of our MedLAM model.

\subsubsection{Multi Scale Similarity (MSS)} 
However, given the inherent variations in anatomical positioning across different individuals, regions sharing the same latent coordinates in various images may still correspond to different anatomical structures. Therefore, we need to further refine the precision of our localization by extracting local pixel-level features from our points of interest. This allows us to pinpoint the most similar feature within the vicinity of the initially localized point, thereby enhancing the overall localization accuracy. This is inspired by the work in Cascade Comparing to Detect (CC2D) \citep{yao2021one}, which ensures that augmented instances of the same image yield highly similar features for the same point, while different points exhibit substantially divergent features.
\par More specifically, as illustrated in Fig. \ref{fig:mss}, the inputs to our MSS process encompass normalized multi-scale feature maps, denoted as $\bm{F}=\{\bm{F}^i,i=0,1,2\}$ and $\bm{F}'=\{\bm{F}^{'i},i=0,1,2\}$, extracted from $\bm{x}$ and $\bm{x}'$, respectively. These multi-scale feature maps consist of three different scales, where the size of each scale is $1/{2^i}$ times the size of the original image, with the largest scale being the same size as the original image.

\par Given a chosen point $\bm{c}_j$ from $\bm{x}$, which has a corresponding point in $\bm{x}'$ is denoted as $\bm{c}'_j$. For every scale $i$ of the multi-scale feature maps, we extract the feature vectors $\bm{f}_{\bm{c}_j}^i$ and $\bm{f}_{\bm{c}'_j}^{'i}$ corresponding to points $\bm{c}_j$ and $\bm{c}'_j$ respectively. Formally, $\bm{f}_{\bm{c}_j}^i=\bm{F}^i(\bm{c}_j)$ and $\bm{f}_{\bm{c}'_j}^{'i}=\bm{F}^{'i}(\bm{c}'_j)$.

\par To compute the similarity between these feature vectors and the corresponding scale feature maps in $\bm{x'}$, we use the cosine similarity metric. The cosine similarity $s$ between two vectors $\bm{a}$ and $\bm{b}$ is given by:

\begin{equation}
s(\bm{a}, \bm{b}) = \frac{\bm{a} \cdot \bm{b}}{||\bm{a}||_2 \cdot ||\bm{b}||_2} .
\end{equation}

\par Since the $\bm{F}^i$ and $\bm{F}'^i$ are both normalized, for each scale $i$, the cosine similarity map is computed by using the feature vector $\bm{f}_{\bm{c}_j}^i$ as a $1\times1\times1$ convolutional kernel that operates on the corresponding feature map from the augmented image. Our optimization objective is to increase the similarity between $\bm{f}_{\bm{c}_j}^i$ and $\bm{f}_{\bm{c}'_j}^{'i}$ relative to other points, effectively enhancing their mutual likeness while diminishing their similarity to the remaining features in the augmented image. To achieve this, we apply the softmax operation across all values in the cosine similarity map, resulting in the final probability map $\bm{S}^i_{\bm{c}_j}$. Correspondingly, the ground truth $\bm{Y}^i_{\bm{c}_j}$ at scale $i$, is set as: 

\begin{equation}
\bm{Y}^i_{\bm{c}_j}(\bm{c}) = 
\begin{cases} 
1, & \text{if}\ \bm{c} = round(\frac{\bm{c}'_j}{2^i}) \\
0, & \text{otherwise}
\end{cases}
\end{equation}
This ensures that all other points are set to 0, while the corresponding position for $\bm{c}'_j$ at scale $i$ is set to 1. We utilize the Cross Entropy (CE) loss to align each $\bm{S}^i_{\bm{c}_j}$ with its corresponding ground truth $\bm{Y}^i_{\bm{c}_j}$:

\begin{equation}
    L_{CE} = \sum_{i} -\bm{Y}^i_{\bm{c}_j}\cdot log(\bm{S}^i_{\bm{c}_j}) - (1-\bm{Y}^i)\cdot log(1-\bm{S}^i_{\bm{c}_j}).
\end{equation}
Eventually, we set the final loss function as:

\begin{equation}
    L_{SSL} = L_{MSE} +L_{CE}.
\end{equation}


\begin{figure}[t!]
    \centering
    \includegraphics[width=1\linewidth]{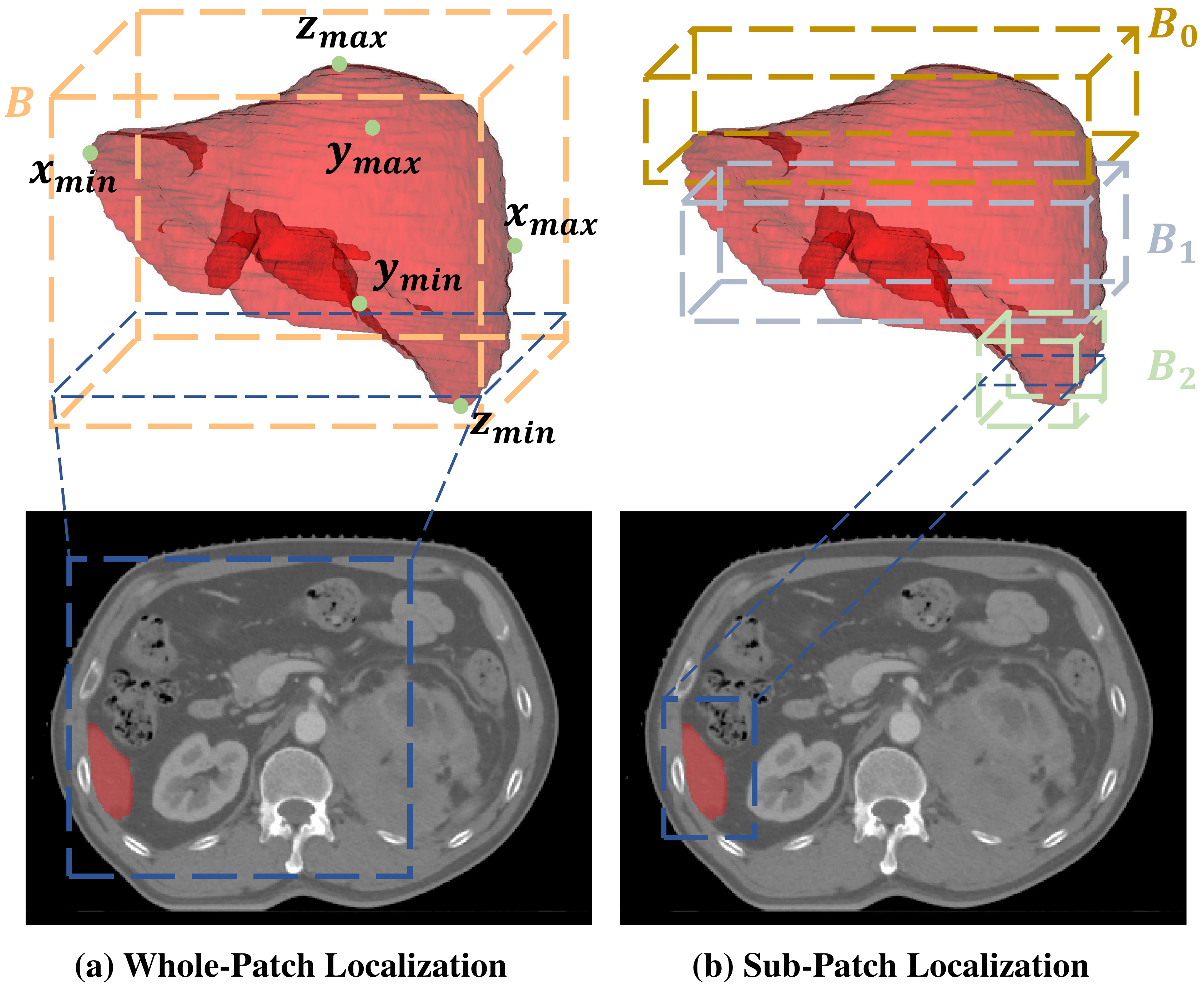}
    \caption{Comparison between Whole-Patch Localization (WPL) and Sub-Patch Localization (SPL) strategies. (a) WPL generates a single minimal 3D bounding box ($\bm{B}$) by identifying the six extreme points of the target anatomical structure. (b) SPL divides the target structure into multiple segments, each with its own localized 3D bounding box ($\bm{B}_i$), resulting in a more accurate representation at each slice.}
    \label{fig:subpatch}
\end{figure} 

\subsection{Inference of MedLSAM}
\label{sec:3.2}
\par  In this section, we will elaborate on the process of utilizing MedLAM to localize any point of interest within scans. Additionally, we further elucidate the mechanism through which MedLAM crafts 3D bbox prompts for the anatomical structures of interest. These 3D prompts are then sliced into 2D bboxes corresponding to each layer of the structure. Each 2D bbox, acting as a prompt for its respective layer, is combined with the corresponding image slice and processed by the SAM model to produce the segmentation results.

\subsubsection{Landmark Localization}
\par As posited in Section \ref{sec:3.1.1}, we assume that the same anatomical structure in different individuals corresponds to the same latent coordinates. This allows us to calculate the latent coordinates of any point within the query volume and subsequently determine its relative distance to the target landmark by computing the difference in their latent coordinates. 
\par As shown in Fig. \ref{fig:inference}, we conceptualize the localization task as moving an agent from a randomly initialized position in the query image toward the target location. A patch is extracted from a random position within the query image, and simultaneously, a support patch is extracted from the support image, centered around the pre-specified landmark. Upon processing these two patches through the MedLAM model, we obtain a 3D offset that represents the estimated relative spatial displacement between the start and target positions and the multi-scale feature vectors of the pre-specified landmark. By updating the agent's location based on this offset, we achieve a coarse localization of the landmark within the query image. 

\par To further refine the landmark localization, we employ the Multi-Scale Similarity (MSS) component of MedLAM. Specifically, we extract a 3D patch of size $D\times W\times H$ centered around the coarsely localized point in the query image. Within the extracted patch, we compute multi-scale feature maps, perform similarity calculations, and aggregate the resulting similarity maps. This allows us to identify the location within the patch that exhibits the highest feature similarity, thereby significantly enhancing the precision of our landmark localization.

\subsubsection{Sub-Patch Localization (SPL)}
\par  As shown in Fig. \ref{fig:subpatch}(a), for a target anatomical structure, we can generate a single minimal 3D bounding box (bbox) $\bm{B}$ that completely encompasses the organ by locating its six extreme points in three dimensions:  $[\bm{z}_{min}, \bm{z}_{max}, \bm{x}_{min}, \bm{x}_{max}, \bm{y}_{min}, \bm{y}_{max}]$. We call this localization approach as \textbf{Whole-Patch Localization (WPL)}. However, the cross-sectional area of this 3D bbox on each layer may significantly exceed the actual cross-sectional area of the organ at that layer, resulting in highly inaccurate bbox prompts.
\par In response, we introduce the \textbf{Sub-Patch Localization (SPL)} strategy. Rather than using a single bounding box, SPL divides the target structure into multiple segments, each segment having its own dedicated 3D bbox, denoted as $\bm{B}_i$. This segmented approach, as visualized in Fig. \ref{fig:subpatch}(b), ensures that each bbox is tailored to more accurately represent the organ's actual cross-sectional area at every layer.
\par Delving into the specifics, for any given target organ, SPL slices the organ into segments at intervals of $n$ mm, resulting in $m$ distinct segments for the organ. Consequently, MedLAM is then tasked with localizing these $m$ segments in the query volume, producing $m$ individual bounding boxes: $[\bm{z}^i_{min}, \bm{z}^i_{max}, \bm{x}^i_{min}, \bm{x}^i_{max}, \bm{y}^i_{min}, \bm{y}^i_{max}]$ for each segment $i$. This refined approach provides a much more accurate spatial representation of the anatomical structure across its entirety.

\subsubsection{SAM Segmentation}
\par After generating the bbox prompts for each slice, we transition to the segmentation stage. For this, we utilize both SAM and MedSAM, a specialized variant of SAM that has been fine-tuned for medical image datasets. Both models serve as the foundation for our segmentation tasks. The versatility of SAM and the domain-specific adaptations of MedSAM help us provide robust segmentation results, thereby adding to the overall efficacy of the MedSLAM system.

\section{Experiments}

\subsection{Dataset}
\par Our MedLAM model is trained on an extensive set of 16 datasets, which collectively comprise a total of 14,012 CT scans. These scans encompass various regions of the human body, providing comprehensive anatomical coverage. An overview of the training datasets is provided in Table~\ref{tab:dataset}. The diverse and abundant training data ensures the robustness and generalizability of our model across different medical imaging contexts.
\par To validate the effectiveness of our approach,  we test \textcolor{black}{MedLAM and MedLSAM on two CT datasets with mask annotation: 1) StructSeg19 Task1 dataset \footnote{https://structseg2019.grand-challenge.org/Home/} for the 22 Head-and-Neck (HaN) organs with 50 scans; 2) the WORD dataset \cite{luo2022word} for the 16 abdomen organs with 120 scans. For MedLSAM, we integrate MedLAM with two segmentation backbones: SAM\cite{kirillov2023segment} and MedSAM\cite{MedSAM}.}

\input{tabel/dataset}

\subsection{Implementation Details}
\label{sec:4.2}
Our model was trained using four NVIDIA GTX 3090 Ti GPUs. We utilized the Adam optimizer~\cite{kingma2014adam} with a batch size of 8, an initial learning rate of $10^{-3}$, and a training duration of 250 epochs. 
\par In terms of pre-processing for MedLAM's training and inference, we rescaled the voxel spacing to [3,3,3] mm and standardized the cropping patch sizes to 64$\times$64$\times$64 pixels. \textcolor{black}{We utilize Segmental Linear Function\cite{lei2021automatic} to normalize the CT image from [-1000, -200, 200, 1500] HU to [0,0.2,0.8,1]}. To ensure that the scanning range was fully covered, we set the parameter $r$ as [1500, 600, 600]. 
\par \textcolor{black}{To separately evaluate the performance of MedLAM and MedLSAM, we focus on tasks including landmark detection, organ localization, and organ segmentation. }  In both the StructSeg HaN and WORD datasets, we start by randomly selecting \textcolor{black}{$k$} scans as support volumes \textcolor{black}{and the remaining as the query volumes}. 
\par \textcolor{black}{For landmark detection, we identify the six extreme points for each organ. Specifically,}  for each target organ, we compute the six extreme coordinates and then average these coordinates and features across the support images. This creates an average representation of latent coordinates and features for each of the organ's extreme points. \textcolor{black}{Once we have these latent coordinates and features, we locate the corresponding points in the query volumes and generate 3D bounding boxes (bbox) for the organ localization task.}
\par \textcolor{black}{For the segmentation tasks, the Whole-Patch Localization (WPL) approach directly utilizes the organ’s 3D bbox, providing a bbox for each slice, as detailed in Sec. \ref{sec:3.2}. In contrast, the Sub-Patch Localization (SPL) method takes a different approach.} Instead of focusing on the extreme coordinates of the organ, we divide each target organ into a number of sub-patches, denoted as $i$. The value of $i$ is determined by taking the minimum span of the target organ across all support volumes and dividing it by the predefined patch distance $n$ mm. This ensures that we have a consistent number of sub-patches for each target organ across all support volumes. For each of these $i$ sub-patches, we crop, compute, and integrate the latent coordinates and features across \textcolor{black}{all} support volumes, offering a more granular representation. As a result, we obtain the latent coordinates and features for 6$i$ points. \textcolor{black}{To account for variations in organ size and shape, we extend each slice’s bbox prompt by 10 pixels in both the $x$ and $y$ directions, ensuring that the entire target structure is adequately captured. }
\par 
\par Upon utilizing SAM and MedSAM, the original images are subjected to a separate preprocessing routine in line with the standard procedures described in the original MedSAM methodology. This includes\textcolor{black}{: 1) cutting the slice intensities to [-500, 1000] HU, and then normalizing its values to [0, 1]; 2) adjusting the slice resolution to 1024 × 1024; 3) duplicating the slice three times and stacking them to form a dimension of 3 × 1024 × 1024.}
\par To guarantee the robustness of our experimental results, we repeat this random selection and averaging process five times. The final performance metrics are obtained by averaging the outcomes of these iterations. \textcolor{black}{We explore different support set sizes $k$ in Sec. \ref{sec:4.4.2}, while maintaining $k$=5 for all other experiments.}

 \subsection{Evaluation Metrics}
 To evaluate the \textcolor{black}{landmark localization and organ localization accuracy}, we apply three widely utilized metrics \citep{li2018fast, humpire2018efficient, xu2019efficient, lei2021contrastive}: \textcolor{black}{the Average Localization Error (LE)}, Wall Distance (WD) and Intersection over Union (IoU). 
 \par \textcolor{black}{Average Localization Error (ALE) measures the average Euclidean distance between the predicted and ground truth points across the six extreme points of each organ. It is defined as:
  \begin{equation}
     ALE= \frac{1}{6}\sum_{i=1}^{6} \sqrt{(x_{p,i} - x_{gt,i})^2 + (y_{p,i} - y_{gt,i})^2 + (z_{p,i} - z_{gt,i})^2}
 \end{equation}
where $x_{p,i}$, $y_{p,i}$, $z_{p,i}$ are the coordinates of the predicted points, and $x_{gt,i}$, $y_{gt,i}$, $z_{gt,i}$ are the coordinates of the corresponding ground truth points for each of the six extreme points. This metric reflects the overall accuracy of the landmark localization across the organ.}
 \par For Wall Distance (WD), it quantifies the average absolute distance difference between the boundaries of the predicted bbox and the ground-truth bbox. Specifically, it is defined as:
 \begin{equation}
     WD = \frac{1}{6}\sum{| d_{p,i} - d_{gt,i} |},
 \end{equation}
where $d_{p,i}$ and $d_{gt,i}$ are the distances of the six boundaries (top, bottom, left, right, front, back) of the predicted and ground-truth bbox, respectively. The pixel spacing is implicitly taken into account in the distance measurement.
 \par The IoU overlap ratio is defined as:
 \begin{equation}
     IoU = \frac{\bm{V}_p \cap \bm{V}_{gt}}{\bm{V}_p \cup \bm{V}_{gt}},
 \end{equation}
where $V_p$ and $V_{gt}$ denote the volume of the predicted bbox and the ground-truth bbox, respectively. The IoU metric (0-1, 0: mismatch; 1: perfectly match) is employed to assess the relative accuracy of the predicted bounding box with respect to the ground truth. 
\par 
\par As for the segmentation metrics, we use Dice Similarity Coefficient (DSC) score 
as evaluation metrics to quantitatively evaluate our method. Dice score  (0-1, 0: mismatch; 1: perfectly match) measures the overlap of the prediction $\bm{A}$ and ground truth $\bm{B}$, and is defined as:
\begin{equation}
    DSC(\bm{A}, \bm{B})=\frac{2|\bm{A}\cap\bm{B}|}{|\bm{A}|+|\bm{B}|}\cdot 100\%
    \label{eq12}
\end{equation}

\subsection{Evaluation of MedLAM}
\subsubsection{Comparison with Other Methods}
\label{sec:4.4.1}

\input{tabel/tab2}

\input{tabel/tab3}

\input{tabel/tab4}

\input{tabel/tab5}

In this section, with the purpose of validating the performance of our universal localization model MedLAM \textcolor{black}{.}. We have conducted a comparative analysis between MedLAM and \textcolor{black}{four well-established methodologies: 1) DetCo~\cite{xie2021detco}, an unsupervised contrastive learning framework for object detection; 2) Mask R-CNN~\cite{he2017mask}, a general fully-supervised detection method; 3) nnDetection~\cite{baumgartner2021nndetection}, which serves as a highly robust fully-supervised medical image detection baseline; 4) MIU-VL~\cite{qin2023medical}, a pioneer work utilizing visual-language foundation model for zero-shot medical image localization.}
\par
\textcolor{black}{For the implementation of DetCo, we apply the identical preprocessing protocol utilized for MedLAM, training it on the same dataset of 14,012 CT scans with a pretrained 3D ResNet50 backbone\cite{feichtenhofer2019slowfast}. During inference, we employ the Multi-Scale Similarity (MSS) process described in Section \ref{sec:4.2}, where we average the features of each landmark across the support volume to accurately localize the most similar point within the query volume. To ensure a fair comparison, we maintained the same support/query split as MedLAM for each experiment.} 
\par
For Mask R-CNN, we utilize the 3D implementation in MedicalDetectionToolkit\cite{jaeger2020retina} \footnote{https://github.com/MIC-DKFZ/medicaldetectiontoolkit}. \textcolor{black}{For the organ extreme point detection task, each extreme point of an organ was treated as a separate class in both Mask R-CNN and nnDetection models. To ensure a thorough and comprehensive performance evaluation, we implemented a five-fold cross-validation approach for both models.} 

\par Regarding MIU-VL, we follow the procedures outlined in the original paper, using GLIP \citep{li2022grounded} as the foundation model for zero-shot detection.  To generate the required text prompts for MIU-VL, we utilized ChatGPT \citep{ouyang2022training} to generate prompts containing descriptions of [COLOR], [SHAPE], [LOCATION], and [CLASS] for different organs.  Since the MIU-VL network is designed to directly handle 2D images, we performed the detection process on 3D medical images slice by slice. Subsequently, we reconstructed the obtained bounding boxes back into the 3D image space. \textcolor{black}{However, we found it challenging to describe organ extreme points effectively using the attributes required by MIU-VL, such as providing specific descriptions of the SHAPE of these points. As a result, we did not include MIU-VL in the landmark detection experiments.}

\par \textcolor{black}{The final experimental results for landmark detection and organ localization are separately presented in Table~\ref{tab:2} and Table~\ref{tab:3}. MedLAM outperforms all other four methods on the StructSeg HaN datasets and achieves performance comparable to Mask R-CNN on the WORD dataset. Specifically, on the StructSeg HaN dataset, MedLAM achieved an average ALE of 4.3 mm and IoU of 48\%, substantially surpassing nnDetection's 23.3 mm, 16.1\% and Mask R-CNN's 24.4mm, 18.4\%. On the WORD dataset, MedLAM's average IoU of 56.6\% surpasses nnDetection by 11.2\%, but is slightly lower than Mask R-CNN. This suggests that MedLAM has an advantage in localizing densely packed small organs.} 

\textcolor{black}{For the self-supervised DetCo method, we have observed that it is only capable of localization within a coarse range. This limitation may be due to DetCo's design for natural images, which relies on the assumption of high contrast— an assumption not typically met in CT images.} We also observed that MIU-VL was only able to delineate the entire body area regardless of the organ-specific prompt, highlighting a substantial gap between foundation models pre-trained on natural images and their applicability in the medical domain. These results also indicate that such natural image foundation models may struggle to provide meaningful localization in low-contrast medical images. In contrast, our MedLAM model, which incorporates strong medical domain knowledge, is capable of localizing any anatomical structure with just a few annotated examples, exceeding the performance of fully supervised models.

To be more specific, in the StructSeg HaN dataset, the organ with the largest average WD (Wall Distance) is the right parotid gland, which does not exceed 4mm. This suggests that MedLAM can perform highly accurate localization for head and neck organs. On the other hand, for the WORD dataset, the abdominal soft tissue organs have blurry boundaries and significant deformations, resulting in larger average WD values. However, the WD for each organ is generally within 15mm.

In summary, MedLAM exhibits robust localization performance and proves to be a versatile foundation model for localizing any type of structure in 3D medical imaging. For smaller organs, although the IoU scores are relatively lower, the subsequent preprocessing step of expanding the 3D bounding box ensures that these organs are adequately captured for segmentation. This makes the model practically applicable across a broad spectrum of organ sizes. This approach provides a pragmatic solution to the challenge of localizing smaller organs with MedLAM, thus achieving a balanced performance across different organ sizes.

\input{tabel/tab6}

\input{tabel/tab7}

\input{tabel/tab8}

\subsubsection{Ablation Study of MedLAM}
\label{sec:4.4.2}
\textcolor{black}{\textbf{Effect of Support Volume Size:}
To explore how the size of the support set impacts the performance of MedLAM, we varied the support volume size $k$ within the WORD dataset. The corresponding IoU scores are presented in Table \ref{tab:4}. Notably, increasing $k$ from 1 to 3 and then from 3 to 5 led to significant improvements in the mean IoU score, from 36.8 to 49.3, and from 49.3 to 56.6, respectively. This trend aligns with MedLAM's underlying hypothesis that the same anatomical parts across different individuals have similar coordinates in a unified anatomical coordinate system. By averaging the coordinates of the same anatomical part across multiple individuals, fluctuations can be minimized, making the average coordinates closer to the target anatomical coordinates in the query volume. However, the performance improvement becomes less pronounced when $k$ increases from 5 to 7. Consequently, we set $k=5$ for all subsequent experiments to maintain the minimum amount of annotation required.}
\par
\textcolor{black}{\textbf{Effect of Each Component:}}
To assess the contributions of individual components within MedLAM, we conduct a series of experiments on the WORD dataset. Specifically, we evaluate MedLAM under the following conditions: 1) using only Unified Anatomical Mapping (UAM) for coarse localization; 2) relying solely on Multi Scale Similarity (MSS) for pinpointing the most similar point across the entire query image; 3) employing RDR for initial coarse localization, followed by MSS with and without features from each scale. The IoU scores obtained from these experiments are presented in Table \ref{tab:5}. As can be observed, when using either RDR or MSS independently, the average IoU is below 50\%, standing at 39.2\% and 34.2\%, respectively. This could be attributed to the following reasons: 1) RDR operates under the assumption of anatomical consistency, which may not hold true if the query volume undergoes significant deformation; 2) MSS focuses on the similarity of local features but lacks an understanding of global anatomical structures, making it susceptible to interference from other areas in the image that are similar to the query point. This results in the identification of false positives, as evidenced by the high variance in the MSS detection results.

However, when initiating with a coarse localization point identified by RDR and subsequently refining it with MSS, the detection accuracy is significantly improved. As shown in Table \ref{tab:5}, the average IoU can be increased from 39.2\% to 56.6\% when using multi-scale features for MSS-based refining. Furthermore, using individual feature maps for $\bm{F}^i$ refining also improves the average IoU to over 50\%, validating the effectiveness of combining global anatomical features with local pixel features.

\textcolor{black}{
\subsubsection{Evaluating MedLAM's Robustness to Anatomical Abnormalities}
Since MedLAM's Unified Anatomical Mapping (UAM) and Multi-Scale Similarity (MSS) are based on the assumption that anatomical structures across different individuals exhibit similar distributions and appearances, this foundation may not hold for patients with substantial organ abnormalities. To assess if MedLAM's ability to localize structures is significantly impacted by pathological versus normal conditions, we tested it on FLARE2023\footnote{https://codalab.lisn.upsaclay.fr/competitions/12239} validation set, which includes organs affected by cancer.}
\par \textcolor{black}{Specifically, the FLARE2023 validation set comprises 100 cases, of which 50 are annotated. Among these, the number of cases with cancer in the liver, left kidney, right kidney, pancreas, and stomach are 26, 27, 18, 9, and 3, respectively. The tumor labels in the validation set were transformed to their corresponding organ labels to assess MedLAM's ability to localize both the organ and the lesion as a single entity. We randomly selected five volumes from the FLARE2023 training set, which have complete organ annotations, to serve as support volumes.  MedLAM localized the aforementioned five types of organs on the annotated validation set, and localization accuracy for organs with and without cancer was calculated separately. The results are presented in Table \ref{tab:6}. It is evident that for organs such as the liver, pancreas, and stomach, MedLAM's localization accuracy remains relatively unaffected. However, the detection accuracy for the left and right kidneys is substantially impacted due to the relatively large size of kidney tumors compared to the overall kidney volume \cite{heller2020state}. Consequently, one limitation of MedLAM is its reduced efficacy in detecting abnormalities in organs where pathological conditions cause significant morphological changes.}

\subsection{Evaluation of MedLSAM}
\subsubsection{\textcolor{black}{Comparison with Other Methods}}
\label{sec:4.5.1}
\begin{figure*}[ht!]
    \centering
    \includegraphics[width=1\linewidth]{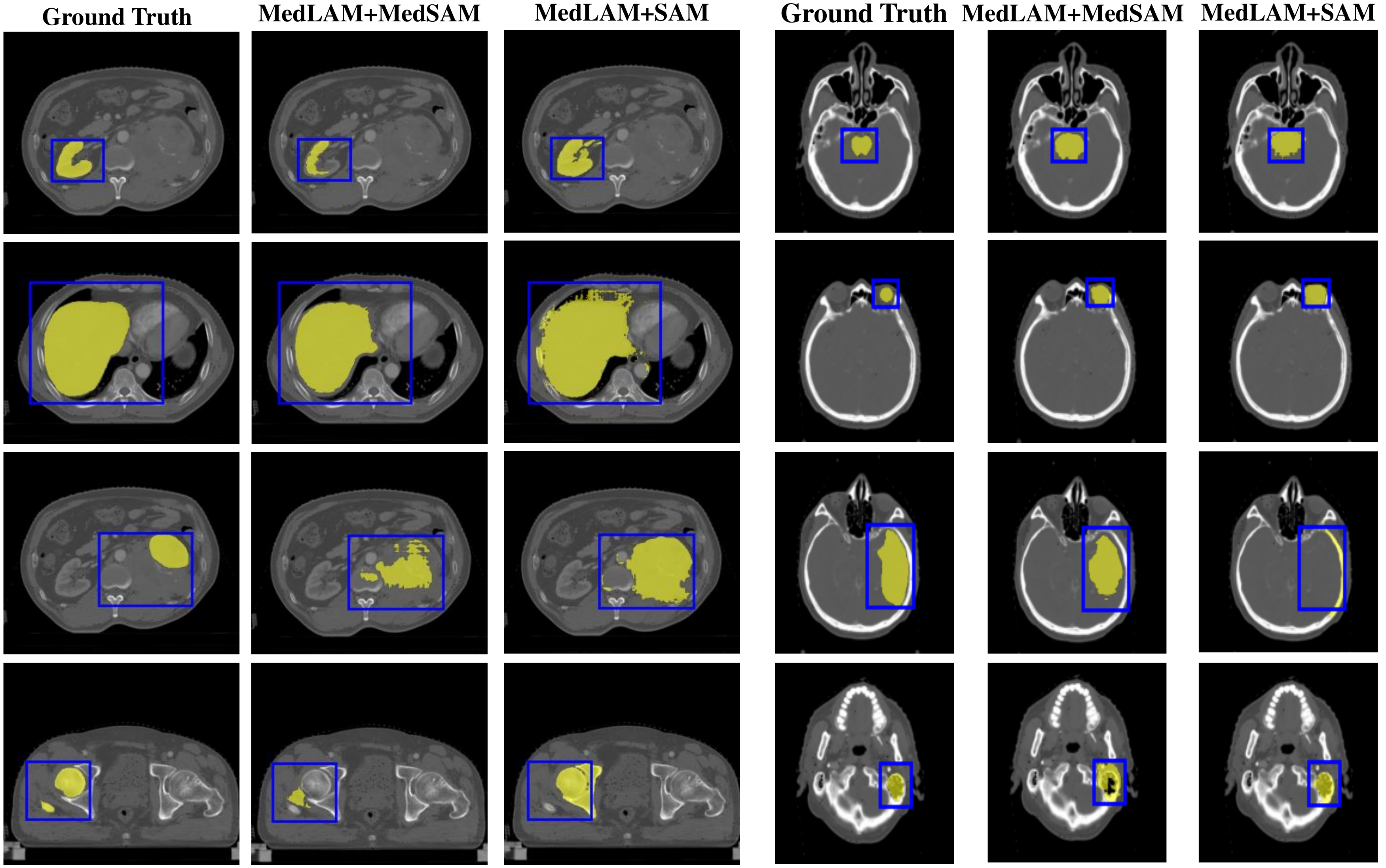}
    \caption{Visualization examples of segmentation results on WORD and StructSeg Head-and-Neck datasets using pre-trained MedSAM and SAM, post landmark localization with MedLAM.}
    \label{fig:seg_result}
\end{figure*}

After validating the localization performance of MedLAM, we proceeded to examine the segmentation performance of the proposed MedLSAM framework. \textcolor{black}{To conduct a comprehensive evaluation of MedLSAM, we compared it against two main types of segmentation methods: 1) Prompt-based segmentation methods; 2) Fully supervised segmentation methods.}

\par More specifically, our experiments with prompt-based methods included:
 \par \noindent\textcolor{black}{1) UniverSeg \cite{butoi2023universeg}, which leverages a universal segmentation model with the example set of image-label pairs to segment unseen tasks without additional training. We employed UniverSeg's default preprocessing for all inputs, ensuring that pixel values are min-max normalized to the range [0, 1] and that the spatial dimensions are set to (H, W) = (128, 128). For each support volume and its associated annotation, we divided each organ into three equal parts and selected the corresponding slices, which were then concatenated to form the support set. The segmentation was performed layer by layer along the axial axis, with the results subsequently concatenated to form the final segmented output;} 
 \par \noindent\textcolor{black}{2) MedLSAM, utilizing the localization result from MedLAM, together with either SAM or MedSAM, to perform segmentation tasks. In each experimental round, the same support volumes and query volumes were used for both UniverSeg and MedLSAM;} 
 \par \noindent\textcolor{black}{3) nnDetection-based prompt, where we utilized nnDetection trained via five-fold cross-validation as described in Sec \ref{sec:4.4.1}. Predicted bounding boxes from each respective validation set were used as prompts for both SAM and MedSAM;} 
 \par \noindent\textcolor{black}{4) Manual prompt - This involves manually annotated bounding boxes that serve as the performance upper bound for SAM and MedSAM. These were simulated based on the ground-truth masks, following the same approach used for generating bboxes in MedSAM. During inference, a bbox prompt was generated from each ground-truth mask, with a random perturbation of 0-20 pixels introduced to mimic the potential inaccuracies that would be present in manually-drawn bboxes.}
\par \textcolor{black}{For the fully supervised methods, we employed nnU-Net \cite{isensee2021nnu}, aiming to establish both lower and upper performance bounds for fully supervised models. For the lower bound model, in each experiment, we used the same support set as employed for UniverSeg and MedLSAM as the training dataset, and tested the performance on the same query set. For the upper bound model, the entire dataset was utilized to conduct five-fold cross-validation, with the results averaged on the validation set to establish the benchmark.
}

\par \textcolor{black}{The results of these experiments are recorded in Table \ref{tab:7}. For the prompt-based methods, it could be observed that MedLSAM demonstrates performance closely approximating that achieved with manual prompts and surpasses the UniSeg and nnDetection-based prompt segmentation results largely for many organs. For instance, in the case of the left and right eye, MedLSAM attains a DSC score of 69.3\% and 69.4\%, while the manual prompt reaches 70.0\% and 69.1\%. However, for certain abdominal organs with substantial shape variability, such as the pancreas, the performance gap between MedLSAM and manual prompts is more significant, with a DSC score difference reaching up to 35\%. Fig. \ref{fig:seg_result} visually presents the results of using MedLSAM for localization and segmentation tasks on both datasets. Overall, MedLSAM performs comparably to manual prompts when segmenting part of the head and neck organs, though it still falls short with large or morphologically complex abdominal organs. Nevertheless, this represents a pioneering and effective effort towards fully automating the use of SAM in medical imaging.}
\par \textcolor{black}{However, our results indicate that the prompt-based segmentation methods still struggle to match the performance of fully supervised segmentation methods. Even nnU-Net, when trained with only five support image-label pairs, can surpass the segmentation results of manual prompts for most abdominal organs. Furthermore, nnU-Net trained on the entire dataset demonstrates a significant advantage. Nonetheless, training fully supervised segmentation models requires substantial time and resources, and these models often lack flexibility, being tailored to specific tasks only. In contrast, medical foundation models like MedSAM represent a more promising approach, aligning more closely with the future of medical scenarios that require complex analysis and interaction\cite{zhang2023challenges,moor2023foundation}. This highlights the urgent need for continued research into medical foundation segmentation models that can offer both efficiency and versatility.}

\subsubsection{Ablation study of SPL}
To validate the effectiveness of Sub-Patch Localization (SPL), we conducted experiments on the StructSeg HaN dataset, comparing the performance of SPL at different sub-patch intervals $n$ mm against the Whole-Patch Localization (WPL) approach. As described in Sec. \ref{sec:4.2}, the WPL method relies on the extreme coordinates of the entire organ for localization. On the other hand, SPL employs a more granular approach by segmenting the organ into multiple sub-patches at intervals of $n$ mm.
\par As shown in Table \ref{tab:8}, it is evident that the SPL approach consistently outperforms WPL. Notably, for organs with significant morphological variations, such as the brain stem and spinal cord, the DSC score improves by more than 7\% and 20\%, respectively.  Moreover, when $n$ is relatively small, SPL enables finer-grained localization by concentrating on smaller, more specific portions of the organ.  This results in bounding boxes that better approximate the organ's actual shape at each slice, thereby enhancing the accuracy of the subsequent segmentation process. 


\subsubsection{\textcolor{black}{Performance on TotalSegmentator}}
\input{tabel/tab9}
\textcolor{black}{
To provide a thorough evaluation of MedLSAM over a broad range of anatomical structures, we selected the TotalSegmentator dataset \cite{wasserthal2023totalsegmentator}. This dataset encompasses 104 anatomical structures annotated across 1204 CT examinations, including 27 organs, 59 bones, 10 muscles, and 8 vessels. Not every scan in TotalSegmentator includes annotations for all 104 structures; therefore, for each organ, we randomly chose five volumes containing that organ to serve as support volumes, while the remaining volumes with that organ were used as query volumes. This method allowed us to assess MedLSAM's performance on each organ individually. We utilize Sub-Patch Localization (SPL) with $n=6$ mm. Similar to Section \ref{sec:4.5.1}, we included a comparison with the performance upper bound using manually annotated bounding boxes as prompts. Only the SAM framework was used for segmentation, as the training set of MedSAM \cite{MedSAM} contains the entire TotalSegmentator dataset.
}
\par \textcolor{black}{
The results are recorded in Table \ref{tab:9}. Compared to organs with highly irregular structures, such as the colon, MedLSAM achieves better segmentation performance on organs with more consistent shapes and structures.
}

\section{Discussion \& Conclusions}
\textcolor{black}{In this study, we introduced MedLAM, a foundational localization model for medical images capable of identifying any anatomical structure within 3D medical images. By integrating it with the SAM model, we developed MedLSAM, an automated segmentation framework designed to reduce the annotation workload in medical image segmentation.}

\textcolor{black}{We validated MedLAM and MedLSAM across two 3D datasets, covering 38 different organs. The results demonstrate that MedLAM, using only a few support samples, can achieve performance comparable to or better than existing fully supervised models. MedLAM’s ability to localize anatomical structures efficiently and accurately underscores its potential for broader applications in medical imaging, including rapid and automatic localization of user-specified landmarks or regions of interest. This makes it particularly suited for scenarios requiring flexible and scalable localization without the need for extensive fine-tuning.}

\par \textcolor{black}{While MedLSAM shows promise in reducing annotation workloads, it still does not match the performance of fully supervised models such as nnU-Net, especially in cases involving organs with significant anatomical abnormalities. This highlights the need for further refinement of prompt-based segmentation methods.}

\par \textcolor{black}{Future work should focus on enhancing MedLAM’s versatility and integrating it with emerging generalist medical AI (GMAI)\cite{moor2023foundation} applications. Doing so could further improve its generalizability and close the performance gap with fully supervised models, while also reducing the resources and time required for medical imaging analysis. MedLAM's strong performance in anatomical structure localization suggests potential for real-time, interactive medical applications, particularly in segmenting and analyzing specific anatomical regions that may involve novel combinations of categories not yet covered by public datasets.}

\section*{Acknowledgments}

\bibliographystyle{model2-names.bst}\biboptions{authoryear}
\bibliography{medlsam.bib}

\end{document}

%% file: tabel/dataset.tex
\begin{table}[t!]
\caption{Detailed information of the 16 CT datasets for MedLAM training.}
\centering
\scalebox{0.75}{
\begin{tabular}{l|cc} 
\hline
Dataset          & Number       & Anatomical Region                    \\ \hline
GLIA \cite{bo2021toward}            & 1338         & HaN                           \\
ACRIN 6685\cite{lowe2019multicenter}       & 260          & HaN                           \\
OPC-Radiomics\cite{kwan2018radiomic}    & 606          & HaN                           \\
Head-Neck-PET-CT\cite{vallieres2017radiomics} & 298          & HaN                          \\
HNSCC \cite{grossberg2018imaging}           & 591          & HaN/Thorax/Abdomen     \\
autoPET \cite{gatidis2022whole}         & 1014         & Whole                       \\
MELA \cite{mela}          & 770          & Thorax                      \\
LIDC-IDRI\cite{armato2011lung}        & 1308         & Thorax                       \\
STOIC2021\cite{revel2021study}        & 2000         & Thorax                       \\
MSD-Lung\cite{antonelli2022medical}         & 95           & Thorax                      \\
CBIS-DDSM\cite{lee2017curated}        & 2620         & Thorax                      \\
AMOS 2022\cite{ji2022amos}        & 500          & Thorax/Abdomen             \\
Kits19\cite{heller2020state}           & 141          & Abdomen                     \\
MSD-Colon\cite{antonelli2022medical}        & 190          & Abdomen                     \\
MSD-Pancreas\cite{antonelli2022medical}     & 281          & Abdomen                      \\
FLARE2022\cite{ma2022fast}        & 2000         & Abdomen                      \\ \hline
Total            & 14,012       &  Whole                           \\ \hline
\end{tabular}}
\label{tab:dataset}
\end{table}

%% file: tabel/tab2.tex
\begin{table}[t!]
\centering
\caption{\textcolor{black}{Comparison of MedLAM with few-shot and fully supervised (FS) localization models on the landmark localization task using the StructSeg Head-and-Neck and WORD datasets. Results are reported in Average Localization Error (ALE, mean$\pm$std mm). "$\dagger$" indicates that the differences between MedLAM and all baseline methods are statistically significant at p $<$ 0.05.}}
\textcolor{black}{
\begin{subtable}{1\linewidth}
\caption{\textcolor{black}{StructSeg Head-and-Neck}}
\scalebox{0.75}{
\begin{tabular}{ccccc}
\toprule[1.5pt]
\multirow{2}*{Organs} & MedLAM & DetCo& Mask R-CNN& nnDetection \\ 
~ & (5-shot) & (5-shot) & (FS) & (FS)\\ \hline
Brain Stem & $\bm{3.5\pm2.3^\dagger}$ & 54.1$\pm$55.8 & 5.2$\pm$0.4 & 15.5$\pm$19.1  \\ \hline
        Eye L & $\bm{3.4\pm1.5}$ & 56.4$\pm$44.1 & 3.5$\pm$0.6 & 11.5$\pm$18.7  \\ \hline
        Eye R & $\bm{3.0\pm1.3^\dagger}$ & 68.6$\pm$48.6 & 3.9$\pm$0.5 & 13.3$\pm$13.9  \\ \hline
        Lens L & $\bm{3.7\pm1.4^\dagger}$ & 97.9$\pm$50.9 & 16.6$\pm$0.7 & 20.5$\pm$34.2  \\ \hline
        Lens R & $\bm{3.1\pm1.7^\dagger}$ & 80.8$\pm$62.8 & 15.4$\pm$0.6 & 25.8$\pm$28.2  \\ \hline
        Opt Nerve L & $\bm{3.6\pm2.3^\dagger}$ & 75.8$\pm$41.6 & 18.3$\pm$38.2 & 18.5$\pm$21.0  \\ \hline
        Opt Nerve R & $\bm{3.8\pm2.1^\dagger}$ & 70.3$\pm$52.1 & 25.1$\pm$57.0 & 31.4$\pm$23.9  \\ \hline
        Opt Chiasma & $\bm{4.1\pm2.0^\dagger}$ & 68.1$\pm$52.6 & 26.3$\pm$34.4 & 14.8$\pm$18.3  \\ \hline
        Temporal Lobes L & $\bm{5.7\pm2.5}$ & 93.6$\pm$51.2 & 5.7$\pm$5.3 & 31.6$\pm$34.0  \\ \hline
        Temporal Lobes R & $\bm{4.6\pm3.2}$ & 64.5$\pm$52.7 & 5.2$\pm$2.2 & 17.9$\pm$22.7  \\ \hline
        Pituitary & $\bm{4.5\pm2.3^\dagger}$ & 89.5$\pm$38.3 & 67.8$\pm$57.3 & 34.7$\pm$27.1  \\ \hline
        Parotid Gland L & 6.6$\pm$2.4 & 64.8$\pm$41.3 & $\bm{6.6\pm1.7}$ & 17.9$\pm$22.7  \\ \hline
        Parotid Gland R & $\bm{6.4\pm3.4^\dagger}$ & 80.3$\pm$52.2 & 32.0$\pm$127.2 & 23.1$\pm$32.3  \\ \hline
        Inner Ear L & $\bm{4.1\pm1.3^\dagger}$ & 57.1$\pm$38.0 & 64.0$\pm$0.0 & 17.4$\pm$21.0  \\ \hline
        Inner Ear R & $\bm{4.7\pm1.4^\dagger}$ & 73.1$\pm$44.4 & 38.8$\pm$0.0 & 29.8$\pm$25.1  \\ \hline
        Mid Ear L & $\bm{5.7\pm3.6^\dagger}$ & 75.0$\pm$55.4 & 22.6$\pm$1.1 & 28.0$\pm$27.2  \\ \hline
        Mid Ear R & $\bm{5.4\pm2.9^\dagger}$ & 74.0$\pm$63.5 & 43.2$\pm$23.2 & 28.3$\pm$26.0  \\ \hline
        TM Joint L & $\bm{4.5\pm1.8^\dagger}$ & 52.7$\pm$46.5 & 47.9$\pm$33.9 & 18.7$\pm$19.5  \\ \hline
        TM Joint R & $\bm{3.9\pm1.5^\dagger}$ & 83.9$\pm$54.3 & 49.3$\pm$33.8 & 33.5$\pm$36.0  \\ \hline
        Spinal Cord & $\bm{4.4\pm3.1^\dagger}$ & 97.6$\pm$40.2 & 27.4$\pm$41.7 & 17.1$\pm$26.3  \\ \hline
        Mandible L & $\bm{3.6\pm2.2^\dagger}$ & 72.1$\pm$46.9 & 7.0$\pm$19.9 & 26.3$\pm$21.5  \\ \hline
        Mandible R & $\bm{3.8\pm2.6^\dagger}$ & 129.2$\pm$45.9 & 5.5$\pm$1.6 & 34.5$\pm$32.3 \\
        \hline
        Average & $\bm{4.3\pm1.2^\dagger}$ & 75.2$\pm$32.7 & 24.4$\pm$14.5 & 23.3$\pm$12.9\\
\bottomrule[1.5pt]
\end{tabular}}
\end{subtable}}%

\hfill

\textcolor{black}{
\begin{subtable}{1\linewidth}
\caption{\textcolor{black}{WORD}}
\scalebox{0.75}{
\begin{tabular}{ccccc}
\toprule[1.5pt]
\multirow{2}*{Organs} & MedLAM & DetCo& Mask R-CNN& nnDetection \\ 
~ & (5-shot) & (5-shot) & (FS) & (FS)\\ \hline
Liver & 20.4$\pm$14.4 & 144.3$\pm$61.5 & $\bm{11.5\pm5.8^\dagger}$ & 38.0$\pm$51.1  \\ \hline
        Spleen & $\bm{9.5\pm7.3}$ & 65.6$\pm$68.8 & 11.9$\pm$3.3 & 22.7$\pm$30.0  \\ \hline
        Kidney L & $\bm{8.8\pm14.3}$ & 61.6$\pm$42.8 & 11.2$\pm$14.5 & 20.4$\pm$36.9  \\ \hline
        Kidney R & $\bm{5.9\pm5.8}$ & 103.9$\pm$50.1 & 7.9$\pm$5.2 & 40.1$\pm$41.4  \\ \hline
        Stomach & 29.3$\pm$20.2 & 71.8$\pm$54.4 & $\bm{14.2\pm4.4^\dagger}$ & 32.7$\pm$36.1  \\ \hline
        Gallbladder & 33.8$\pm$15.6 & 137.0$\pm$75.7 & $\bm{8.8\pm7.1^\dagger}$ & 38.1$\pm$43.2  \\ \hline
        Esophagus & 12.2$\pm$9.2 & 85.8$\pm$44.2 & $\bm{6.2\pm2.8^\dagger}$ & 27.4$\pm$32.1  \\ \hline
        Pancreas & 23.3$\pm$13.5 & 64.0$\pm$58.2 & $\bm{13.5\pm3.5^\dagger}$ & 37.7$\pm$47.1  \\ \hline
        Duodenum & 19.6$\pm$15.6 & 81.7$\pm$58.4 & $\bm{13.5\pm5.5^\dagger}$ & 39.5$\pm$42.6  \\ \hline
        Colon & 28.7$\pm$20.1 & 75.1$\pm$77.8 & $\bm{21.5\pm11.3^\dagger}$ & 43.9$\pm$47.0  \\ \hline
        Intestine & 28.9$\pm$13.2 & 97.1$\pm$65.4 & $\bm{18.9\pm7.9^\dagger}$ & 39.2$\pm$44.9  \\ \hline
        Adrenal & 12.6$\pm$9.5 & 82.5$\pm$51.5 & $\bm{10.8\pm7.7^\dagger}$ & 24.1$\pm$46.6  \\ \hline
        Rectum & 13.2$\pm$8.1 & 104.6$\pm$53.4 & $\bm{9.5\pm16.5^\dagger}$ & 33.4$\pm$31.2  \\ \hline
        Bladder & 14.2$\pm$9.5 & 109.1$\pm$82.0 & $\bm{6.9\pm3.0^\dagger}$ & 34.3$\pm$41.8  \\ \hline
        Head of Femur L & 10.6$\pm$27.3 & 97.5$\pm$48.8 & $\bm{5.3\pm2.8^\dagger}$ & 20.8$\pm$38.6 \\ \hline
        Head of Femur R & 9.7$\pm$11.7 & 88.8$\pm$72.7 & $\bm{6.2\pm6.1^\dagger}$ & 46.7$\pm$68.3  \\ \hline
        Average & 17.5$\pm$10.8 & 92.2$\pm$45.4 & $\bm{11.1\pm4.2^\dagger}$ & 33.6$\pm$30.2\\
\bottomrule[1.5pt]
\end{tabular}}
\end{subtable}
}
\label{tab:2}
\end{table}

%% file: tabel/tab3.tex
\begin{table*}[t!]
\centering
\caption{Comparison of MedLAM with few-shot, fully supervised (FS), and zero-shot localization models on the organ detection task using the StructSeg Head-and-Neck and WORD datasets. Results are reported in IoU and Wall Distance (WD). "$\dagger$" indicates that the differences between MedLAM and all baseline methods are statistically significant at p $<$ 0.05.}
\textcolor{black}{
\begin{subtable}{1\textwidth}
\centering
\caption{\textcolor{black}{StructSeg Head-and-Neck}}
\scalebox{0.75}{
\begin{tabular}{@{}ccccccccccc@{}}
\toprule[1.5pt]
\multirow{3}*{Organs}  & \multicolumn{5}{c}{IoU $\uparrow$ (mean $\pm$ std \%)} & \multicolumn{5}{c}{WD $\downarrow$ (mean $\pm$ std mm)}\\ \cmidrule(lr){2-6} \cmidrule(lr){7-11} 
~ & MedLAM & DetCo& Mask R-CNN& nnDetection & MIU-VL & MedLAM & DetCo& Mask R-CNN& nnDetection & MIU-VL \\ 
~ & (5-shot) & (5-shot) & (FS) & (FS)& (zero-shot) & (5-shot) & (5-shot) & (FS) & (FS)& (zero-shot)\\ \hline
Brain Stem&70.3 $\pm$ 8.0&44.8 $\pm$ 1.5&64.5 $\pm$ 5.2&$\bm{75.7\pm6.4}^\dagger$&0.1 $\pm$ 0.0&$\bm{2.4\pm1.4}^\dagger$&36.3 $\pm$ 31.8&3.1 $\pm$ 0.3&8.5 $\pm$ 11.9&223.0 $\pm$ 74.3 \\
Eye L&$\bm{66.2\pm11.5}^\dagger$&14.9 $\pm$ 45.4&62.7 $\pm$ 7.2&29.0 $\pm$ 35.8&0.0 $\pm$ 0.0&$\bm{1.7\pm1.0}^\dagger$&30.1 $\pm$ 30.5&2.2 $\pm$ 0.4&7.2 $\pm$ 13.5&162.8 $\pm$ 73.9 \\
Eye R&63.2 $\pm$ 11.2&25.9 $\pm$ 27.1&$\bm{63.3\pm}$8.0&42.7 $\pm$ 28.4&0.0 $\pm$ 0.0&$\bm{1.8\pm1.1}$&36.5 $\pm$ 28.5&2.1 $\pm$ 0.3&8.0 $\pm$ 10.6&163.4 $\pm$ 61.1 \\
Lens L&$\bm{11.5\pm11.0}$&5.9 $\pm$ 11.6&1.6 $\pm$ 0.8&10.2 $\pm$ 16.0&0.0 $\pm$ 0.0&$\bm{1.9\pm1.1}^\dagger$&49.7 $\pm$ 32.5&10.4 $\pm$ 0.5&13.8 $\pm$ 19.5&220.2 $\pm$ 89.7 \\
Lens R&17.9 $\pm$ 14.5&5.5 $\pm$ 16.6&1.0 $\pm$ 1.4&$\bm{24.3\pm22.2}$&0.0 $\pm$ 0.0&$\bm{1.8\pm1.2}^\dagger$&55.9 $\pm$ 37.4&9.2 $\pm$ 0.4&16.8 $\pm$ 19.9&219.1 $\pm$ 75.6 \\
Opt Nerve L&$\bm{29.7\pm27.0}^\dagger$&8.3 $\pm$ 12.2&7.7 $\pm$ 15.3&15.4 $\pm$ 15.6&0.0 $\pm$ 0.0&$\bm{2.0\pm1.3}^\dagger$&40.5 $\pm$ 33.6&11.4 $\pm$ 27.4&9.5 $\pm$ 14.9&173.0 $\pm$ 56.3 \\
Opt Nerve R&$\bm{27.6\pm25.4}^\dagger$&12.0 $\pm$ 15.5&6.2 $\pm$ 17.1&23.6 $\pm$ 17.1&0.0 $\pm$ 0.0&$\bm{2.1\pm1.3}$&49.0 $\pm$ 29.9&14.0 $\pm$ 33.5&16.2 $\pm$ 18.5&175.7 $\pm$ 47.8 \\
Opt Chiasma&9.7 $\pm$ 19.4&2.9 $\pm$ 18.2&10.2 $\pm$ 4.7&$\bm{32.4\pm9.4}^\dagger$&0.5 $\pm$ 0.0&$\bm{2.8\pm1.5}^\dagger$&38.7 $\pm$ 29.4&16.0 $\pm$ 24.9&9.0 $\pm$ 13.0&177.0 $\pm$ 25.2 \\
Temporal Lobes L&70.9 $\pm$ 8.7&26.2 $\pm$ 46.5&$\bm{73.0\pm13.8}$&45.3 $\pm$ 37.1&0.5 $\pm$ 0.1&$\bm{3.2\pm1.8}$&48.5 $\pm$ 39.6&3.4 $\pm$ 3.0&17.0 $\pm$ 21.0&163.3 $\pm$ 33.9 \\
Temporal Lobes R&$\bm{72.8\pm10.3}$&15.2 $\pm$ 42.8&71.8 $\pm$ 11.7&26.8 $\pm$ 33.8&0.0 $\pm$ 0.2&$\bm{3.1\pm1.8}$&39.2 $\pm$ 30.2&3.4 $\pm$ 1.7&9.5 $\pm$ 14.8&189.0 $\pm$ 64.9 \\
Pituitary&8.9 $\pm$ 18.5&5.3 $\pm$ 15.8&4.8 $\pm$ 12.6&$\bm{25.6\pm16.8}^\dagger$&0.3 $\pm$ 0.0&$\bm{2.7\pm1.6}^\dagger$&46.9 $\pm$ 31.9&37.1 $\pm$ 39.3&24.4 $\pm$ 19.5&229.2 $\pm$ 84.7 \\
Parotid Gland L&$\bm{63.1\pm9.6}$&17.1 $\pm$ 27.6&61.0 $\pm$ 5.0&31.3 $\pm$ 32.9&0.3 $\pm$ 0.1&$\bm{3.7\pm1.9}$&44.8 $\pm$ 32.3&4.1 $\pm$ 1.0&11.8 $\pm$ 15.5&162.7 $\pm$ 40.2 \\
Parotid Gland R&$\bm{62.7\pm10.3}^\dagger$&21.2 $\pm$ 39.0&19.5 $\pm$ 24.2&38.7 $\pm$ 32.0&0.0 $\pm$ 0.1&$\bm{4.0\pm2.1}^\dagger$&51.4 $\pm$ 33.3&19.1 $\pm$ 73.9&13.6 $\pm$ 20.8&170.1 $\pm$ 44.6 \\
Inner Ear L&$\bm{35.6\pm8.2}$&19.4 $\pm$ 15.9&15.2 $\pm$ 12.2&34.1 $\pm$ 25.2&0.0 $\pm$ 0.0&$\bm{2.5\pm1.0}^\dagger$&38.9 $\pm$ 29.3&32.7 $\pm$ 36.4&10.3 $\pm$ 12.2&182.3 $\pm$ 34.8 \\
Inner Ear R&$\bm{33.7\pm9.4}$&19.8 $\pm$ 29.5&0.1 $\pm$ 0.4&32.1 $\pm$ 27.7&0.1 $\pm$ 0.0&$\bm{2.7\pm1.1}^\dagger$&44.7 $\pm$ 36.3&24.5 $\pm$ 26.3&15.3 $\pm$ 16.9&185.4 $\pm$ 28.0 \\
Mid Ear L&$\bm{56.5\pm12.9}^\dagger$&19.0 $\pm$ 28.0&3.9 $\pm$ 8.6&31.0 $\pm$ 32.9&0.1 $\pm$ 0.1&$\bm{3.4\pm2.0}^\dagger$&48.6 $\pm$ 32.1&12.1 $\pm$ 0.7&14.0 $\pm$ 15.7&193.6 $\pm$ 72.0 \\
Mid Ear R&$\bm{56.5\pm12.8}$&26.5 $\pm$ 32.7&6.1 $\pm$ 4.7&51.1 $\pm$ 27.5&0.0 $\pm$ 0.1&$\bm{3.5\pm2.1}^\dagger$&47.0 $\pm$ 35.7&22.6 $\pm$ 14.4&15.0 $\pm$ 17.1&204.6 $\pm$ 77.5 \\
TM Joint L&$\bm{40.7\pm15.1}^\dagger$&14.5 $\pm$ 20.1&8.2 $\pm$ 23.4&26.8 $\pm$ 22.3&0.0 $\pm$ 0.0&$\bm{2.4\pm1.3}^\dagger$&35.0 $\pm$ 33.5&25.5 $\pm$ 22.7&11.3 $\pm$ 13.9&188.8 $\pm$ 57.6 \\
TM Joint R&$\bm{38.7\pm15.3}^\dagger$&17.1 $\pm$ 21.1&10.6 $\pm$ 19.4&26.0 $\pm$ 18.4&0.5 $\pm$ 0.0&$\bm{2.3\pm1.2}^\dagger$&48.0 $\pm$ 32.3&24.8 $\pm$ 27.2&21.4 $\pm$ 21.4&200.2 $\pm$ 64.7 \\
Spinal Cord&$\bm{57.7\pm16.7}^\dagger$&15.8 $\pm$ 22.5&6.3 $\pm$ 11.5&10.4 $\pm$ 14.8&0.9 $\pm$ 0.2&$\bm{2.7\pm1.9}^\dagger$&50.3 $\pm$ 29.1&14.9 $\pm$ 26.1&11.3 $\pm$ 15.1&224.5 $\pm$ 120.1 \\
Mandible L&$\bm{80.9\pm6.2}^\dagger$&18.2 $\pm$ 33.5&72.5 $\pm$ 13.0&36.1 $\pm$ 37.3&1.0 $\pm$ 0.3&$\bm{2.5\pm1.6}^\dagger$&42.5 $\pm$ 31.1&4.9 $\pm$ 15.0&13.8 $\pm$ 17.7&164.2 $\pm$ 73.1 \\
Mandible R&$\bm{80.8\pm5.9}^\dagger$&17.2 $\pm$ 30.3&79.8 $\pm$ 6.6&34.3 $\pm$ 40.0&1.0 $\pm$ 0.4&$\bm{2.6\pm1.5}^\dagger$&66.9 $\pm$ 34.6&2.9 $\pm$ 1.1&21.4 $\pm$ 20.4&155.2 $\pm$ 59.6 \\ \hline
Average&$\bm{48.0\pm13.1}^\dagger$&19.6 $\pm$ 24.8&29.6 $\pm$ 10.3&31.9 $\pm$ 25.0&0.2 $\pm$ 0.1&$\bm{2.6\pm1.5}^\dagger$&45.0 $\pm$ 32.5&13.7 $\pm$ 14.3&13.6 $\pm$ 16.5&187.6 $\pm$ 61.8 \\
\bottomrule[1.5pt]
\end{tabular}}
\end{subtable}}

\vspace{0.2cm}

\textcolor{black}{
\begin{subtable}{1\textwidth}
\centering
\caption{\textcolor{black}{WORD}}
\scalebox{0.75}{
\begin{tabular}{@{}ccccccccccc@{}}
\toprule[1.5pt]
\multirow{3}*{Organs}  & \multicolumn{5}{c}{IoU $\uparrow$ (mean $\pm$ std \%)} & \multicolumn{5}{c}{WD $\downarrow$ (mean $\pm$ std mm)}\\ \cmidrule(lr){2-6} \cmidrule(lr){7-11} 
~ & MedLAM & DetCo& Mask R-CNN& nnDetection & MIU-VL & MedLAM & DetCo& Mask R-CNN& nnDetection & MIU-VL \\ 
~ & (5-shot) & (5-shot) & (FS) & (FS)& (zero-shot) & (5-shot) & (5-shot) & (FS) & (FS)& (zero-shot)\\ \hline
Liver&73.0 $\pm$ 11.6&30.5 $\pm$ 13.9&$\bm{79.8\pm7.4}^\dagger$&52.7 $\pm$ 10.2&10.6 $\pm$ 2.1&10.5$\pm$9.1&73.8  $\pm$  48.3&$\bm{7.1 \pm 3.4}^\dagger$&23.8 $\pm$ 31.5&80.4 $\pm$ 37.2 \\
Spleen&$\bm{70.9\pm13.3}$&23.2 $\pm$ 13.1&69.0 $\pm$ 9.6&38.2 $\pm$ 17.5&1.7 $\pm$ 1.0&$\bm{5.7\pm5.9}$&37.9  $\pm$  40.2&6.0 $\pm$ 2.2&14.6 $\pm$ 21.4&142.5 $\pm$ 49.8 \\
Kidney L&71.0 $\pm$ 15.9&44.0 $\pm$ 30.5&$\bm{74.0\pm16.6}$&69.6 $\pm$ 27.2&0.9 $\pm$ 0.5&$\bm{5.0\pm8.4}$&39.4  $\pm$  33.2&5.7 $\pm$ 10.1&13.9 $\pm$ 22.1&164.3 $\pm$ 69.2 \\
Kidney R&$\bm{76.0\pm13.8}$&49.3 $\pm$ 8.3&74.3 $\pm$ 11.2&74.2 $\pm$ 11.5&0.8 $\pm$ 0.2&$\bm{3.6\pm4.6}$&61.4  $\pm$  40.5&4.1 $\pm$ 3.0&20.8 $\pm$ 28.6&166.1 $\pm$ 81.3 \\
Stomach&49.1 $\pm$ 14.3&22.5 $\pm$ 30.4&$\bm{58.4\pm10.6}^\dagger$&36.3 $\pm$ 23.3&4.7 $\pm$ 2.1&17.7 $\pm$ 12.8&49.3  $\pm$  38.0&8.8 $\pm$ 3.7&$\bm{17.0\pm25.1}$&102.8 $\pm$ 25.1 \\
Gallbladder&12.0 $\pm$ 11.0&15.2 $\pm$ 23.1&29.3 $\pm$ 25.0&$\bm{29.3\pm18.2}^\dagger$&0.2 $\pm$ 0.2&16.9$\pm$10.3&71.4  $\pm$  48.1&$\bm{5.1 \pm 4.7}^\dagger$&21.2 $\pm$ 29.6&153.2 $\pm$ 38.5\\
Esophagus&44.2 $\pm$ 17.9&34.9 $\pm$ 6.7&48.9 $\pm$ 18.9&$\bm{58.5\pm13.2}^\dagger$&0.3 $\pm$ 0.2&6.8 $\pm$ 6.7&47.3  $\pm$  29.6&$\bm{4.1 \pm 1.9}^\dagger$&16.8 $\pm$ 18.9&153.8 $\pm$ 22.7\\
Pancreas&44.1 $\pm$ 17.3&27.8 $\pm$ 27.0&65.0 $\pm$ 10.5&$\bm{53.4\pm17.6}^\dagger$&1.6 $\pm$ 0.6&12.2 $\pm$ 8.8&42.0  $\pm$  44.0& $\bm{7.0 \pm 2.6}^\dagger$&19.1 $\pm$ 28.5&135.2 $\pm$ 33.8\\
Duodenum&44.5 $\pm$ 17.5&21.6 $\pm$ 14.8&$\bm{58.5\pm13.4}^\dagger$&42.0 $\pm$ 16.4&1.2 $\pm$ 0.4&12.7 $\pm$ 9.1&56.6  $\pm$  43.9& $\bm{8.1 \pm 3.6}^\dagger$&20.5 $\pm$ 26.1&139.8 $\pm$ 31.0\\
Colon&67.0 $\pm$ 13.0&16.8 $\pm$ 16.3&$\bm{73.4\pm12.3}^\dagger$&28.5 $\pm$ 10.9&17.9 $\pm$ 4.4&15.6 $\pm$ 11.8&49.0  $\pm$  55.1&$\bm{12.3 \pm 8.1}^\dagger$&22.8 $\pm$ 35.8&83.3 $\pm$ 33.6\\
Intestine&62.6 $\pm$ 11.0&12.7 $\pm$ 8.2&$\bm{70.9\pm10.8}^\dagger$&24.6 $\pm$ 10.7&12.2 $\pm$ 4.6&15.4 $\pm$ 9.8&50.7  $\pm$  41.0& $\bm{11.3 \pm 4.7}^\dagger$ &20.5 $\pm$ 29.5&106.0 $\pm$ 53.1\\
Adrenal&40.4 $\pm$ 17.7&33.5 $\pm$ 16.7&55.2 $\pm$ 21.9&$\bm{57.3\pm12.7}^\dagger$&0.4 $\pm$ 0.2&7.6 $\pm$ 5.5&52.3  $\pm$  38.7& $\bm{6.2 \pm 5.9}$ &16.1 $\pm$ 26.6&160.2 $\pm$ 58.0\\
Rectum&49.3 $\pm$ 15.3&33.3 $\pm$ 20.0&14.8 $\pm$ 23.0&$\bm{62.6\pm16.9}^\dagger$&0.6 $\pm$ 0.3&8.6 $\pm$ 6.1&54.7  $\pm$  37.1& $\bm{5.8 \pm 10.6}^\dagger$ &17.4 $\pm$ 24.6&203.1 $\pm$ 146.3\\
Bladder&55.4 $\pm$ 16.9&41.8 $\pm$ 21.2&57.4 $\pm$ 30.5&$\bm{63.0\pm17.0}^\dagger$&1.2 $\pm$ 0.7&8.1$\pm$7.6&65.1  $\pm$  52.4& $\bm{3.7 \pm 2.3}^\dagger$ &24.1 $\pm$ 33.2&167.7 $\pm$ 96.9\\
Head of Femur L&$\bm{76.7\pm16.7}^\dagger$&12.9 $\pm$ 17.7&70.5 $\pm$ 23.0&20.0 $\pm$ 26.8&1.6 $\pm$ 2.4&5.3 $\pm$ 16.0&55.8  $\pm$  38.0& $\bm{3.3 \pm 1.9}^\dagger$&14.5 $\pm$ 23.0&175.1 $\pm$ 127.4\\
Head of Femur R&$\bm{69.4\pm14.4}^\dagger$&28.1 $\pm$ 37.2&60.2 $\pm$ 29.4&47.7 $\pm$ 34.0&1.5 $\pm$ 0.8&5.9 $\pm$ 6.9&52.1  $\pm$  59.1& $\bm{4.1 \pm 4.1}^\dagger$ &23.8 $\pm$ 41.5&163.4 $\pm$ 117.5\\ \hline
Average&56.6 $\pm$ 14.8&28.0 $\pm$ 19.1&$\bm{59.8\pm17.2}$&47.4 $\pm$ 17.8&3.6 $\pm$ 1.3&9.9 $\pm$ 8.7&53.9  $\pm$  42.9& $\bm{6.5 \pm 4.6}^\dagger$ &19.2 $\pm$ 27.9&143.6 $\pm$ 63.8 \\
\bottomrule[1.5pt]
\end{tabular}}
\end{subtable}
}
\label{tab:3}
\end{table*}

%% file: tabel/tab4.tex
\begin{table}[ht!]
\centering
\textcolor{black}{
\caption{\textcolor{black}{Impact of the support volume size $k$ of MedLAM on organ detection: IoU score $\uparrow$ (mean $\pm$ std \%) on the WORD dataset.}}
\centering
\scalebox{0.8}{
\begin{tabular}{@{}ccccc@{}}
\toprule[1.5pt]
\multirow{2}*{Organs} & \multicolumn{4}{c}{$k$} \\ \cmidrule(lr){2-5}
~ & 1 & 3 & 5 & 7 \\ \hline
 Liver & 57.9 $\pm$ 12.7 & 69.0 $\pm$ 11.4 & 73.0 $\pm$ 11.6 & \textbf{76.1 $\pm$ 10.7} \\ 
 Spleen & 53.7 $\pm$ 14.3 & 50.9 $\pm$ 19.9 & 70.9 $\pm$ 13.3 & \textbf{72.2 $\pm$ 12.6} \\ 
 Kidney L & 51.0 $\pm$ 25.1 & 67.6 $\pm$ 16.8 & 71.0 $\pm$ 15.9 & \textbf{71.0 $\pm$ 15.7} \\ 
 Kidney R & 43.1 $\pm$ 21.0 & 66.9 $\pm$ 14.4 & 76.0 $\pm$ 13.8 & \textbf{76.1 $\pm$ 11.5} \\ 
 Stomach & 37.9 $\pm$ 15.4 & 43.6 $\pm$ 13.1 & 49.1 $\pm$ 14.3 & \textbf{52.9 $\pm$ 14.6} \\ 
 Gallbladder & 06.0 $\pm$ 7.4 & 12.3 $\pm$ 11.9 & 12.0 $\pm$ 11.0 & \textbf{13.8 $\pm$ 17.1} \\ 
 Esophagus & 35.2 $\pm$ 19.3 & 32.9 $\pm$ 15.1 & 44.2 $\pm$ 17.9 & \textbf{45.8 $\pm$ 17.2} \\ 
 Pancreas & 20.5 $\pm$ 10.5 & 29.4 $\pm$ 14.7 & 44.1 $\pm$ 17.3 & \textbf{52.8 $\pm$ 17.3} \\ 
 Duodenum & 23.7 $\pm$ 14.7 & 31.5 $\pm$ 16.1 & 44.5 $\pm$ 17.5 & \textbf{45.4 $\pm$ 16.7} \\ 
 Colon & 47.3 $\pm$ 10.1 & 60.4 $\pm$ 12.1 & 67.0 $\pm$ 13.0 & \textbf{68.2 $\pm$ 12.0} \\ 
 Intestine & 47.7 $\pm$ 13.1 & 56.1 $\pm$ 11.6 & 62.6 $\pm$ 11.0 & \textbf{64.0 $\pm$ 12.2} \\ 
 Adrenal & 23.3 $\pm$ 13.8 & 30.7 $\pm$ 18.4 & 40.4 $\pm$ 17.7 & \textbf{41.1 $\pm$ 17.4} \\ 
 Rectum & 18.6 $\pm$ 9.6 & 46.8 $\pm$ 15.0 & 49.3 $\pm$ 15.3 & \textbf{50.3 $\pm$ 14.5} \\ 
 Bladder & 16.0 $\pm$ 12.4 & 46.9 $\pm$ 15.6 & \textbf{55.4 $\pm$ 16.9} & 53.1 $\pm$ 17.3 \\ 
 Head of Femur L & 57.4 $\pm$ 12.7 & 72.6 $\pm$ 13.9 & \textbf{76.7 $\pm$ 16.7} & 76.3 $\pm$ 14.7 \\ 
 Head of Femur R & 50.2 $\pm$ 10.3 & 71.9 $\pm$ 14.6 & 69.4 $\pm$ 14.4 & \textbf{70.4 $\pm$ 14.3} \\ \hline
 Average & 36.8 $\pm$ 13.9 & 49.3 $\pm$ 14.7 & 56.6 $\pm$ 14.8 & \textbf{58.1 $\pm$ 14.7} \\
\bottomrule[1.5pt]
\end{tabular}}
\label{tab:4}}
\end{table}

%% file: tabel/tab5.tex
\begin{table*}[ht!]
\centering
\caption{Ablation study of Unified Anatomical Mapping (UAM) and Multi Scale Similarity (MSS) in MedLAM for organ detection: IoU score $\uparrow$ (mean $\pm$ std \%) on the WORD dataset. \textcolor{black}{"$\dagger$" means the differences between UAM+MSS and using UAM or MSS alone are significant at p $<$ 0.05.}}
\centering
\scalebox{0.84}{
\begin{tabular}{@{}cccccccccc@{}}
\toprule[1.5pt]
\multirow{3}*{Organs} & UAM & \XSolidBrush & UAM & UAM & UAM & UAM & UAM & UAM & UAM\\
~ & + & + & + & + & + & + & + & + & + \\
~ & \XSolidBrush & $\bm{F}^2$ + $\bm{F}^1$ + $\bm{F}^0$ & $\bm{F}^2$ & $\bm{F}^1$ & $\bm{F}^0$ & $\bm{F}^1$ + $\bm{F}^0$ & $\bm{F}^2$ + $\bm{F}^0$ & $\bm{F}^2$ + $\bm{F}^1$ & $\bm{F}^2$ + $\bm{F}^1$ + $\bm{F}^0$ \\ \hline
Liver& 57.1 $\pm$ 10.5 & 39.7 $\pm$ 27.4 & 76.0 $\pm$ 8.3 & 76.7 $\pm$ 7.9 & 71.5 $\pm$ 10.1 & 75.0 $\pm$ 09.1 & $\bm{78.5\pm7.7}$ & 76.6 $\pm$ 7.7 & 73.0 $\pm$ 11.6$^\dagger$ \\
Spleen& 40.1 $\pm$ 7.7 & 49.3 $\pm$ 34.2 & 63.0 $\pm$ 15.7 & 72.1 $\pm$ 16.0 & 69.1 $\pm$ 21.7 & $\bm{74.7\pm15.2}$ & 74.4 $\pm$ 14.1 & 72.6 $\pm$ 12.9 & 70.9 $\pm$ 13.3$^\dagger$ \\
Kidney L& 45.9 $\pm$ 11.4 & 39.9 $\pm$ 30.7 & 60.5 $\pm$ 10.4 & 58.9 $\pm$ 19.4 & 52.1 $\pm$ 23.4 & 54.5 $\pm$ 22.9 & 64.8 $\pm$ 16.4 & 63.7 $\pm$ 14.8 & $\bm{71.0\pm15.9}$$^\dagger$ \\
Kidney R& 40.4 $\pm$ 9.3 & 40.6 $\pm$ 34.6 & 62.7 $\pm$ 12.1 & 68.2 $\pm$ 14.4 & 54.4 $\pm$ 27.8 & 73.0 $\pm$ 13.1 & $\bm{76.3\pm8.9}$ & 70.6 $\pm$ 9.4 & 76.0 $\pm$ 13.8$^\dagger$ \\
Stomach& 38.7 $\pm$ 12.2 & 25.4 $\pm$ 21.9 & 47.4 $\pm$ 15.7 & 43.4 $\pm$ 15.6 & 46.4 $\pm$ 16.7 & 47.2 $\pm$ 17.6 & 45.9 $\pm$ 12.9 & 47.3 $\pm$ 14.8 & $\bm{49.1\pm14.3}$$^\dagger$ \\
Gallbladder& 10.2 $\pm$ 12.9 & 4.1 $\pm$ 9.5 & 10.1 $\pm$ 9.5 & 7.4 $\pm$ 7.1 & 7.3 $\pm$ 8.1 & 9.3 $\pm$ 13.1 & 9.2 $\pm$ 6.8 & 8.9 $\pm$ 7.0 & $\bm{12.0\pm11.0}$ \\
Esophagus& 22.0 $\pm$ 9.2 & 1.5 $\pm$ 3.4 & 41.0 $\pm$ 15.6 & 39.5 $\pm$ 18.4 & 46.4 $\pm$ 20.4 & 43.6 $\pm$ 19.2 & 44.2 $\pm$ 20.5 & 41.0 $\pm$ 18.5 & $\bm{44.2\pm17.9}$$^\dagger$ \\
Pancreas& 42.9 $\pm$ 13.5 & 13.6 $\pm$ 15.5 & 39.8 $\pm$ 11.5 & 44.0 $\pm$ 18.8 & 34.1 $\pm$ 21.6 & 42.5 $\pm$ 17.4 & $\bm{45.0\pm18.6}$ & 42.8 $\pm$ 16.2 & 44.1 $\pm$ 17.3 \\
Duodenum& 32.2 $\pm$ 10.5 & 31.4 $\pm$ 17.8 & 45.8 $\pm$ 16.2 & 35.8 $\pm$ 20.3 & 31.6 $\pm$ 18.2 & 39.3 $\pm$ 17.6 & 42.8 $\pm$ 15.4 & $\bm{51.0\pm13.4}$ & 44.5 $\pm$ 17.5$^\dagger$ \\
Colon& 60.9 $\pm$ 13.4 & 42.8 $\pm$ 25.6 & 63.7 $\pm$ 12.9 & 65.3 $\pm$ 12.1 & 61.0 $\pm$ 11.1 & 65.2 $\pm$ 11.8 & 64.5 $\pm$ 11.7 & 63.2 $\pm$ 12.5 & $\bm{67.0\pm13.0}$$^\dagger$ \\
Intestine& 58.7 $\pm$ 11.6 & 27.7 $\pm$ 21.1 & 63.1 $\pm$ 13.0 & 61.8 $\pm$ 16.8 & 62.0 $\pm$ 17.0 & 61.9 $\pm$ 17.6 & $\bm{65.2\pm13.1}$ & 63.9 $\pm$ 14.0 & 62.6 $\pm$ 11.0 \\
Adrenal& 29.8 $\pm$ 17.0 & 10.5 $\pm$ 15.6 & 36.8 $\pm$ 16.1 & 30.5 $\pm$ 18.8 & 32.2 $\pm$ 19.7 & 35.0 $\pm$ 19.7 & 38.1 $\pm$ 15.7 & 38.6 $\pm$ 13.5 & $\bm{40.4\pm17.7}$$^\dagger$ \\
Rectum& 34.6 $\pm$ 10.5 & 47.9 $\pm$ 18.2 & 49.1 $\pm$ 12.4 & 47.5 $\pm$ 11.6 & 32.1 $\pm$ 19.4 & 47.9 $\pm$ 12.7 & 51.8 $\pm$ 12.7 & $\bm{53.3\pm13.7}$ & 49.3 $\pm$ 15.3$^\dagger$ \\
Bladder& 33.9 $\pm$ 15.2 & 45.8 $\pm$ 23.0 & 49.5 $\pm$ 18.1 & 55.3 $\pm$ 21.1 & 49.6 $\pm$ 20.1 & 52.3 $\pm$ 22.2 & 50.8 $\pm$ 20.4 & $\bm{55.4\pm16.9}$ & 55.4 $\pm$ 16.9$^\dagger$ \\
Head of Femur L& 43.7 $\pm$ 10.8 & 71.9 $\pm$ 22.4 & 71.8 $\pm$ 8.5 & 80.2 $\pm$ 8.7 & $\bm{84.1\pm8.9}$ & 83.1 $\pm$ 8.9 & 81.6 $\pm$ 9.1 & 77.9 $\pm$ 8.5 & 76.7 $\pm$ 16.7$^\dagger$ \\
Head of Femur R& 36.2 $\pm$ 10.9 & 55.3 $\pm$ 29.4 & 63.8 $\pm$ 13.3 & 70.9 $\pm$ 12.5 & $\bm{75.9\pm13.7}$ & 74.0 $\pm$ 14.0 & 70.2 $\pm$ 14.2 & 68.3 $\pm$ 13.9 & 69.4 $\pm$ 14.4$^\dagger$ \\
\hline
Average& 39.2 $\pm$ 11.7 & 34.2 $\pm$ 21.9 & 52.8 $\pm$ 13.1 & 53.6 $\pm$ 15.0 & 50.6 $\pm$ 17.4 & 54.9 $\pm$ 15.8 & 56.5 $\pm$ 13.6 & 55.9 $\pm$ 13.2 & $\bm{56.6\pm14.8}$$^\dagger$ \\
\bottomrule[1.5pt]
\end{tabular}}
\label{tab:5}
\end{table*}

%% file: tabel/tab6.tex
\begin{table}[ht!]
\centering
\textcolor{black}{
\caption{\textcolor{black}{Performance of MedLAM in localizing normal and abnormal organs in the FLARE2023 validation dataset. "$\dagger$" means the differences between normal and abnormal organs are significant at p $<$ 0.05.}}
\centering
\scalebox{0.9}{
\begin{tabular}{@{}ccccc@{}}
\toprule[1.5pt]
\multirow{2}*{Organs} & \multicolumn{2}{c}{IoU $\uparrow$ (mean $\pm$ std \%)} & \multicolumn{2}{c}{WD $\downarrow$ (mean $\pm$ std mm)} \\ \cmidrule(lr){2-3} \cmidrule(lr){4-5} 
~ & Normal & Abnormal & Normal & Abnormal \\ \hline
Liver & \textbf{71.2 $\pm$ 11.1} & 69.2 $\pm$ 9.7 & 12.0 $\pm$  5.8 & \textbf{12.0 $\pm$ 4.7} \\
Kidney L & $\bm{68.0 \pm 21.4}^\dagger$ & 55.8 $\pm$ 24.7 & \textbf{7.1 $\pm$  6.6$^\dagger$} & 14.9 $\pm$ 19.1 \\ 
Kidney R & \textbf{74.9 $\pm$ 20.0} & 67.7 $\pm$ 19.7 & \textbf{4.7 $\pm$  6.0} & 6.8 $\pm$ 6.1 \\ 
Pancreas & 35.9 $\pm$ 15.0 & \textbf{36.8 $\pm$ 7.1} & 17.9 $\pm$  7.0 & \textbf{14.1 $\pm$ 2.9} \\ 
Stomach & 45.9 $\pm$ 15.6 & \textbf{64.5 $\pm$ 11.9 $^\dagger$} & 18.7 $\pm$  7.9 & \textbf{12.3 $\pm$ 4.8$^\dagger$} \\ \hline
Average & \textbf{59.2 $\pm$ 16.6} & 58.8 $\pm$ 14.6 & 12.1 $\pm$ 6.7 & \textbf{12.0 $\pm$ 7.5} \\
\bottomrule[1.5pt]
\end{tabular}}
\label{tab:6}}
\end{table}

%% file: tabel/tab7.tex
\begin{table*}[t!]
    \centering
    \caption{\textcolor{black}{Comparison of segmentation performance between prompt-based and fully-supervised segmentation methods on the StructSeg Head-and-Neck and the WORD datasets using DSC scores $\uparrow$ (mean $\pm$ std \%). In the table, '5-shot' indicates that nnU-Net was trained using the same five support sets as used for MedLAM and UniverSeg, while 'full' refers to conducting five-fold cross-validation using the entire dataset.}}
    \textcolor{black}{
    \begin{subtable}{1\textwidth}
    \centering
    \caption{StructSeg Head-and-Neck}
    \scalebox{0.85}{
    \begin{tabular}{cccccccccc}
    \toprule[1.5pt]
    \multirow{3}*{Organs} & \multicolumn{7}{c}{\textcolor{black}{Prompt-Based}} & \multicolumn{2}{c}{\textcolor{black}{Fully Supervised}}\\ \cmidrule(lr){2-8} \cmidrule(lr){9-10}
    ~ & \multirow{2}*{UniSeg} & \multicolumn{2}{c}{MedLAM} & \multicolumn{2}{c}{nnDetection} & \multicolumn{2}{c}{Manual Prompt} & \multicolumn{2}{c}{nnU-Net} \\ \cmidrule(lr){3-8} \cmidrule(lr){9-10}
    ~ & ~ & SAM & MedSAM & SAM & MedSAM & SAM & MedSAM & 5-shot & full\\\hline
    Brain Stem& 11.9 $\pm$ 3.5 &63.5 $\pm$ 6.3&73.3 $\pm$ 5.0&55.3 $\pm$ 7.7&68.8 $\pm$ 7.5&66.2 $\pm$ 2.7&$\bm{75.9\pm2.7}$&84.4 $\pm$ 1.1&87.5 $\pm$ 0.8 \\
    Eye L&7.5 $\pm$ 3.3&64.5 $\pm$ 6.5&69.3 $\pm$ 5.9&36.9 $\pm$ 29.0&40.1 $\pm$ 31.8&65.7 $\pm$ 5.6&$\bm{70.0\pm4.6}$&52.6 $\pm$ 9.8&74.3 $\pm$ 10.4\\
    Eye R&9.7 $\pm$ 3.7&67.3 $\pm$ 5.8&$\bm{69.4\pm5.5}$&25.4 $\pm$ 32.9&25.4 $\pm$ 33.0&68.9 $\pm$ 4.8&69.1 $\pm$ 5.2&53.1 $\pm$ 7.7&75.2 $\pm$ 11.3\\
    Lens L&0.0 $\pm$ 0.0&15.9 $\pm$ 7.8&16.0 $\pm$ 6.7&28.1 $\pm$ 22.5&12.5 $\pm$ 10.6&$\bm{22.7\pm5.6}$&19.2 $\pm$ 3.5&40.3 $\pm$ 6.9&65.6 $\pm$ 10.2\\
    Lens R&0.0 $\pm$ 0.0&13.8 $\pm$ 8.8&14.0 $\pm$ 5.5&20.2 $\pm$ 21.2&9.1 $\pm$ 9.8&$\bm{22.8\pm7.7}$&17.3 $\pm$ 4.6&38.5 $\pm$ 7.4&63.7 $\pm$ 13.2\\
    Opt Nerve L&0.1 $\pm$ 0.1&23.7 $\pm$ 6.1&23.5 $\pm$ 6.2&15.2 $\pm$ 13.4&12.8 $\pm$ 11.5&32.6 $\pm$ 9.3&$\bm{34.7\pm5.4}$&39.0 $\pm$ 6.3&57.7 $\pm$ 6.8\\
    Opt Nerve R&0.2 $\pm$ 0.3&27.8 $\pm$ 10.2&26.3 $\pm$ 6.6&16.8 $\pm$ 16.5&14.0 $\pm$ 13.3&28.5 $\pm$ 6.7&$\bm{32.4\pm5.2}$&40.2 $\pm$ 7.8&62.2 $\pm$ 7.8\\
    Opt Chiasma&0.0 $\pm$ 0.0&11.4 $\pm$ 10.6&14.4 $\pm$ 11.4&28.0 $\pm$ 10.3&24.0 $\pm$ 05.4&39.8 $\pm$ 10.2&$\bm{39.8\pm8.1}$&38.4 $\pm$ 8.5&51.4 $\pm$ 6.2\\
    Temporal Lobes L&16.4 $\pm$ 4.2&28.2 $\pm$ 15.2&78.3 $\pm$ 3.5&16.5 $\pm$ 22.1&35.6 $\pm$ 35.6&36.8 $\pm$ 16.6&$\bm{83.5\pm1.9}$&52.5 $\pm$ 8.8&73.7 $\pm$ 10.5\\
    Temporal Lobes R&26.5 $\pm$ 5.6&24.1 $\pm$ 17.4&78.0 $\pm$ 4.3&30.7 $\pm$ 31.1&41.2 $\pm$ 33.3&30.7 $\pm$ 17.8&$\bm{84.4\pm1.4}$&51.7 $\pm$ 11.7&76.7 $\pm$ 6.2\\
    Pituitary&0.0 $\pm$ 0.0&12.5 $\pm$ 10.7&10.2 $\pm$ 9.2&$\bm{43.1 \pm 21.4}$&29.9 $\pm$ 16.5&36.6$\pm$16.2&27.5 $\pm$ 12.5&50.0 $\pm$ 8.6&58.3 $\pm$ 5.9\\
    Parotid Gland L&13.9 $\pm$ 4.3&15.5 $\pm$ 11.9&59.6 $\pm$ 6.5&4.2 $\pm$ 4.3&31.4 $\pm$ 25.1&29.5 $\pm$ 7.7&$\bm{64.4\pm3.4}$&42.8 $\pm$ 6.4&63.3 $\pm$ 9.8\\
    Parotid Gland R&25.7 $\pm$ 7.2&17.2 $\pm$ 9.9&57.1 $\pm$ 6.8&5.1 $\pm$ 7.4&17.9 $\pm$ 23.1&29.6 $\pm$ 6.5&$\bm{64.9\pm5.0}$&48.0 $\pm$ 7.5&70.7 $\pm$ 3.4\\
    Inner Ear L&2.8 $\pm$ 1.6&40.4 $\pm$ 11.8&42.3 $\pm$ 9.9&49.7 $\pm$ 19.6&43.3 $\pm$ 16.9&59.7 $\pm$ 11.7&$\bm{62.1\pm9.7}$&53.3 $\pm$ 12.5&77.9 $\pm$ 5.6\\
    Inner Ear R&0.4 $\pm$ 0.6&48.9 $\pm$ 9.5&45.9 $\pm$ 11.2&49.2 $\pm$ 20.0&43.3 $\pm$ 17.3&$\bm{67.3\pm10.4}$&66.9 $\pm$ 6.1&52.3 $\pm$ 9.1&77.9 $\pm$ 6.4\\
    Mid Ear L&5.4 $\pm$ 3.4&64.6 $\pm$ 14.3&59.7 $\pm$ 9.6&51.4 $\pm$ 22.0&45.6 $\pm$ 18.1&$\bm{71.0\pm11.7}$&67.5 $\pm$ 6.3&44.7 $\pm$ 9.5&72.2 $\pm$ 6.4\\
    Mid Ear R&6.4 $\pm$ 2.4&64.7 $\pm$ 13.1&59.3 $\pm$ 11.2&31.8 $\pm$ 32.0&24.0 $\pm$ 24.1&$\bm{73.3\pm7.7}$&65.0 $\pm$ 8.9&45.4 $\pm$ 7.9&71.8 $\pm$ 7.4\\
    TM Joint L&7.6 $\pm$ 5.8&38.3 $\pm$ 10.1&39.0 $\pm$ 10.9&38.7 $\pm$ 24.2&36.1 $\pm$ 23.1&59.7 $\pm$ 12.8&$\bm{61.2\pm4.9}$&42.6 $\pm$ 8.1&61.4 $\pm$ 6.6\\
    TM Joint R&3.8 $\pm$ 4.6&41.5 $\pm$ 10.0&38.3 $\pm$ 9.5&36.9 $\pm$ 22.0&34.2 $\pm$ 20.4&59.2 $\pm$ 17.8&$\bm{60.0\pm7.7}$&40.6 $\pm$ 9.4&64.9 $\pm$ 5.6\\
    Spinal Cord&$\bm{43.7\pm10.3}$&27.9 $\pm$ 8.3&34.7 $\pm$ 6.9&21.1 $\pm$ 9.3&21.3 $\pm$ 8.0&38.0 $\pm$ 7.7&40.3 $\pm$ 6.1&82.5 $\pm$ 1.2&84.0 $\pm$ 0.6\\
    Mandible L&31.9 $\pm$ 4.2&78.0 $\pm$ 4.9&66.7 $\pm$ 6.1&26.0 $\pm$ 15.8&9.1 $\pm$ 12.2&$\bm{86.0\pm2.1}$&76.8 $\pm$ 4.7&52.6 $\pm$ 8.3&74.6 $\pm$ 8.2\\
    Mandible R&32.8 $\pm$ 6.3&71.4 $\pm$ 4.0&66.0 $\pm$ 4.7&35.9 $\pm$ 20.7&24.5 $\pm$ 15.5&$\bm{81.2\pm2.7}$&75.0 $\pm$ 4.5&54.2 $\pm$ 8.9&77.3 $\pm$ 7.1\\ \hline
    Average&11.2 $\pm$ 3.3&39.1 $\pm$ 9.7&47.3 $\pm$ 7.4&30.3 $\pm$ 19.3&29.3 $\pm$ 18.7&50.3 $\pm$ 9.2&$\bm{57.2\pm5.6}$&50.0 $\pm$ 7.9&70.2 $\pm$ 7.2\\
    \bottomrule[1.5pt]
    \end{tabular}}
    \end{subtable}
    }
    
    \vspace{0.2cm}
    
    \textcolor{black}{
    \begin{subtable}{1\textwidth}
    \centering
    \caption{WORD}
    \scalebox{0.85}{
    \begin{tabular}{cccccccccc}
    \toprule[1.5pt]
    \multirow{3}*{Organs} & \multicolumn{7}{c}{Prompt-Based} & \multicolumn{2}{c}{Fully Supervised}\\ \cmidrule(lr){2-8} \cmidrule(lr){9-10}
    ~ & \multirow{2}*{UniSeg} & \multicolumn{2}{c}{MedLAM} & \multicolumn{2}{c}{nnDetection} & \multicolumn{2}{c}{Manual Prompt} & \multicolumn{2}{c}{nnU-Net} \\ \cmidrule(lr){3-8} \cmidrule(lr){9-10}
    ~ & ~ & SAM & MedSAM & SAM & MedSAM & SAM & MedSAM & 5-shot & full\\\hline
Liver&46.2 $\pm$ 8.6&66.0 $\pm$ 10.1&23.8 $\pm$ 8.6&54.7 $\pm$ 14.2&34.3 $\pm$ 10.8&$\bm{84.2\pm6.3}$&46.6 $\pm$ 14.7&94.3 $\pm$ 0.6&96.3 $\pm$ 0.2 \\
Spleen&28.5 $\pm$ 10.1&61.7 $\pm$ 14.4&36.3 $\pm$ 13.2&54.3 $\pm$ 20.9&35.3 $\pm$ 12.6&$\bm{85.3\pm5.1}$&65.0 $\pm$ 6.3&90.9 $\pm$ 1.7&95.7 $\pm$ 0.4\\
Kidney L&21.7 $\pm$ 6.0&82.1 $\pm$ 16.0&70.7 $\pm$ 18.8&79.1 $\pm$ 18.7&67.5 $\pm$ 17.4&$\bm{92.1\pm1.6}$&84.1 $\pm$ 5.2&83.4 $\pm$ 3.2&94.7 $\pm$ 1.2\\
Kidney R&25.5 $\pm$ 6.8&88.3 $\pm$ 4.8&77.3 $\pm$ 6.2&83.9 $\pm$ 3.8&69.4 $\pm$ 06.5&$\bm{92.9\pm1.6}$&86.4 $\pm$ 3.1&86.0 $\pm$ 2.4&95.2 $\pm$ 0.6\\
Stomach&17.4 $\pm$ 8.4&44.6 $\pm$ 15.0&37.2 $\pm$ 14.7&33.3 $\pm$ 15.7&37.6 $\pm$ 12.9&77.1 $\pm$ 10.6&$\bm{80.3\pm4.7}$&83.2 $\pm$ 2.7&93.1 $\pm$ 1.3\\
Gallbladder&4.9 $\pm$ 3.9&13.1 $\pm$ 17.3&10.2 $\pm$ 13.4&36.9 $\pm$ 21.3&34.2 $\pm$ 18.7&$\bm{72.7\pm11.1}$&68.8 $\pm$ 7.8&54.7 $\pm$ 4.6&77.0 $\pm$ 4.6\\
Esophagus&10.3 $\pm$ 9.6&36.6 $\pm$ 14.7&27.8 $\pm$ 12.9&30.5 $\pm$ 11.9&34.2 $\pm$ 09.5&$\bm{67.0\pm6.4}$&63.1 $\pm$ 7.7&72.0 $\pm$ 0.7&81.5 $\pm$ 1.5\\
Pancreas&8.3 $\pm$ 3.1&29.7 $\pm$ 14.1&21.4 $\pm$ 9.7&26.7 $\pm$ 13.1&14.3 $\pm$ 6.3&$\bm{64.4\pm7.7}$&46.9 $\pm$ 11.6&72.0 $\pm$ 1.3&84.7 $\pm$ 1.6\\
Duodenum&7.5 $\pm$ 3.2&26.0 $\pm$ 11.4&21.1 $\pm$ 7.0&31.7 $\pm$ 11.0&24.7 $\pm$ 8.5&$\bm{54.1\pm13.2}$&51.0 $\pm$ 11.9&53.7 $\pm$ 2.9&77.3 $\pm$ 4.9\\
Colon&25.7 $\pm$ 6.1&25.6 $\pm$ 9.6&26.6 $\pm$ 8.9&23.2 $\pm$ 9.2&22.0 $\pm$ 9.1&41.8 $\pm$ 6.8&$\bm{44.1\pm8.8}$&74.7 $\pm$ 2.1&86.1 $\pm$ 1.2\\
Intestine&29.8 $\pm$ 4.3&37.5 $\pm$ 7.6&34.1 $\pm$ 8.4&28.0 $\pm$ 11.8&23.5 $\pm$ 08.0&$\bm{61.4\pm6.9}$&52.5 $\pm$ 8.0&77.6 $\pm$ 2.1&88.0 $\pm$ 1.1\\
Adrenal&2.0 $\pm$ 2.2&3.3 $\pm$ 3.9&10.0 $\pm$ 6.2&3.9 $\pm$ 8.1&12.4 $\pm$ 08.7&17.4 $\pm$ 8.8&$\bm{26.5\pm5.9}$&58.3 $\pm$ 5.3&72.8 $\pm$ 3.8\\
Rectum&13.5 $\pm$ 7.2&50.1 $\pm$ 17.9&46.0 $\pm$ 18.7&53.2 $\pm$ 13.6&47.5 $\pm$ 13.5&75.5 $\pm$ 4.0&$\bm{80.0\pm3.6}$&70.2 $\pm$ 1.5&80.3 $\pm$ 2.5\\
Bladder&28.7 $\pm$ 18.0&65.3 $\pm$ 26.6&59.1 $\pm$ 23.1&65.7 $\pm$ 17.8&60.2 $\pm$ 14.8&$\bm{83.0\pm15.5}$&82.9 $\pm$ 8.0&84.0 $\pm$ 1.0&91.8 $\pm$ 4.0\\
Head of Femur L&26.9 $\pm$ 6.7&81.7 $\pm$ 3.9&71.5 $\pm$ 4.1&29.4 $\pm$ 36.5&21.1 $\pm$ 26.2&$\bm{90.5\pm2.7}$&80.3 $\pm$ 2.8&43.1 $\pm$ 10.1&31.2 $\pm$ 30.1\\
Head of Femur R&26.6 $\pm$ 5.3&80.1 $\pm$ 3.3&74.3 $\pm$ 4.4&46.2 $\pm$ 37.9&36.6 $\pm$ 30.2&$\bm{89.1\pm3.4}$&83.2 $\pm$ 2.7&47.8 $\pm$ 13.2&40.8 $\pm$ 28.3\\ \hline
Average&20.2 $\pm$ 6.8&49.5 $\pm$ 11.9&40.5 $\pm$ 11.1&42.5 $\pm$ 16.6&35.9 $\pm$ 13.4&$\bm{71.8\pm7.0}$&65.1 $\pm$ 7.0&71.6 $\pm$ 3.5&80.4 $\pm$ 5.5\\
    \bottomrule[1.5pt]
    \end{tabular}}
    \end{subtable}
    }
    \label{tab:7}
    \end{table*}

%% file: tabel/tab8.tex
\begin{table*}[t!]
\centering
\caption{Comparison of DSC scores $\uparrow$ (mean  $\pm$  std\%) for MedLSAM using different localization strategies on the StructSeg Head-and-Neck dataset: Whole-Patch Localization (WPL) and Sub-Patch Localization (SPL) with varying slice intervals $n$. \textcolor{black}{"$\dagger$" means the differences between SPL and WPL are significant at p $<$ 0.05.}}
\centering
\scalebox{0.92}{
\begin{tabular}{ccccccccc}
\toprule[1.5pt]
\multirow{3}*{Organs} & \multicolumn{4}{c}{SAM} & \multicolumn{4}{c}{MedSAM} \\ \cmidrule(lr){2-5} \cmidrule(lr){6-9}
~ & \multicolumn{3}{c}{SPL} & WPL & \multicolumn{3}{c}{SPL} & WPL  \\ \cmidrule(lr){2-4} \cmidrule(lr){5-5} \cmidrule(lr){6-8} \cmidrule(lr){9-9}
~ & $n=6$ mm & $n=15$ mm & $n=30$ mm & \XSolidBrush & $n=6$ mm & $n=15$ mm & $n=30$ mm & \XSolidBrush  \\ \hline
Brain Stem & $\bm{63.5\pm6.3}$$^\dagger$& 60.1 $\pm$ 6.7&51.5 $\pm$ 6.0& 50.8 $\pm$ 6.2&$\bm{73.3\pm5.0}$$^\dagger$&  71.9 $\pm$ 4.8&   67.4 $\pm$ 4.2&  66.6 $\pm$ 4.4\\
Eye L &$\bm{64.5\pm6.5}$$^\dagger$& 61.2 $\pm$ 6.7&61.2 $\pm$ 6.7& 61.2 $\pm$ 6.7&  $\bm{69.3\pm5.9}$$^\dagger$&  66.4 $\pm$ 5.7&   66.4 $\pm$ 5.7&  66.4 $\pm$ 5.7\\
Eye R &$\bm{67.3\pm5.8}$$^\dagger$& 64.5 $\pm$ 6.6&64.5 $\pm$ 6.6& 64.5 $\pm$ 6.6&  $\bm{69.4\pm5.5}$$^\dagger$&  67.8 $\pm$ 6.0&   67.8 $\pm$ 6.0&  67.8 $\pm$ 6.0\\
Lens L &$\bm{15.9\pm7.8}$& 15.7 $\pm$ 7.7&15.7 $\pm$ 7.7& 15.7 $\pm$ 7.7&  $\bm{16.0\pm6.7}$&  15.8 $\pm$ 6.7&   15.8 $\pm$ 6.7&  15.8 $\pm$ 6.7\\
Lens R & 13.8 $\pm$ 8.8&$\bm{14.1\pm8.6}$&13.5 $\pm$ 8.7& 13.5 $\pm$ 8.7&  $\bm{14.0\pm5.5}$&  $\bm{14.0\pm5.6}$&   13.7 $\pm$ 5.5&  13.7 $\pm$ 5.5\\
Opt Nerve L & 23.7 $\pm$ 6.1&$\bm{23.9\pm5.9}$&23.7 $\pm$ 6.1& 23.7 $\pm$ 6.1&  23.5 $\pm$ 6.2&  $\bm{23.6\pm6.2}$&   23.5 $\pm$ 6.2&  23.5 $\pm$ 6.2\\
Opt Nerve R &$\bm{27.8\pm10.2}$&$\bm{27.8\pm10.2}$&$\bm{27.8\pm10.2}$&$\bm{27.8\pm10.2}$&  $\bm{26.3\pm6.6}$&  $\bm{26.3\pm6.6}$&   $\bm{26.3\pm6.6}$&  $\bm{26.3\pm6.6}$\\
Opt Chiasma &$\bm{11.4\pm10.6}$&$\bm{11.4\pm10.6}$& $\bm{11.4\pm10.6}$& $\bm{11.4\pm10.6}$& $\bm{14.4\pm11.4}$& $\bm{14.4\pm11.4}$&  $\bm{14.4\pm11.4}$& $\bm{14.4\pm11.4}$\\
Temporal Lobes L &$\bm{28.2\pm15.2}$&21.7 $\pm$ 14.8& 25.7 $\pm$ 13.4& 25.7 $\pm$ 13.4&  $\bm{78.3\pm3.5}$$^\dagger$&  73.1 $\pm$ 3.5&   71.0 $\pm$ 3.6&  71.0 $\pm$ 3.6\\
Temporal Lobes R &$\bm{24.1\pm17.4}$$^\dagger$&18.2 $\pm$ 17.7& 18.5 $\pm$ 16.4& 18.5 $\pm$ 16.4&  $\bm{78.0\pm4.3}$$^\dagger$&  75.9 $\pm$ 3.1&   71.0 $\pm$ 3.5&  71.0 $\pm$ 3.5\\
Pituitary &12.5 $\pm$ 10.7&$\bm{12.6\pm10.7}$&$\bm{12.6\pm10.7}$&$\bm{12.6\pm10.7}$&  10.2 $\pm$ 9.2&  $\bm{10.3\pm9.4}$&   $\bm{10.3\pm9.4}$&  $\bm{10.3\pm9.4}$\\
Parotid Gland L &$\bm{15.5\pm11.9}$$^\dagger$& 10.4 $\pm$ 9.6& 5.5 $\pm$ 6.6& 5.5 $\pm$ 6.6&  $\bm{59.6\pm6.5}$$^\dagger$&  56.4 $\pm$ 6.6&   48.1 $\pm$ 7.3&  48.1 $\pm$ 7.3\\
Parotid Gland R &$\bm{17.2\pm9.9}$$^\dagger$&13.1 $\pm$ 10.5& 7.7 $\pm$ 7.7& 7.7 $\pm$ 7.7&  $\bm{57.1\pm6.8}$$^\dagger$ &  54.6 $\pm$ 6.8&   45.3 $\pm$ 7.8&  45.3 $\pm$ 7.8\\
Inner Ear L &$\bm{40.4\pm11.8}$&$\bm{40.4\pm11.8}$&$\bm{40.4\pm11.8}$&$\bm{40.4\pm11.8}$&  $\bm{42.3\pm9.9}$&  $\bm{42.3\pm9.9}$&  $\bm{42.3\pm9.9}$ & $\bm{42.3\pm9.9}$ \\
Inner Ear R &$\bm{48.9\pm9.5}$&$\bm{48.9\pm9.5}$&$\bm{48.9\pm9.5}$&$\bm{48.9\pm9.5}$& $\bm{45.9\pm11.2}$ & $\bm{45.9\pm11.2}$ & $\bm{45.9\pm11.2}$ & $\bm{45.9\pm11.2}$ \\
Mid Ear L &$\bm{64.6\pm14.3}$&62.3 $\pm$ 14.7& 62.3 $\pm$ 14.7& 62.3 $\pm$ 14.7& $\bm{59.7\pm9.6}$$^\dagger$ & 55.8 $\pm$ 10.0&  55.7 $\pm$ 10.0& 55.7 $\pm$ 10.0\\
Mid Ear R &$\bm{64.7\pm13.1}$&62.4 $\pm$ 13.1& 62.4 $\pm$ 13.0& 62.4 $\pm$ 13.0& $\bm{59.3\pm11.2}$$^\dagger$ & 53.6 $\pm$ 12.1&  53.4 $\pm$ 11.9& 53.4 $\pm$ 11.9\\
TM Joint L &38.3 $\pm$ 10.1&38.3 $\pm$ 10.1& 38.3 $\pm$ 10.1& 38.3 $\pm$ 10.1& $\bm{39.0\pm10.9}$ & $\bm{39.0\pm10.9}$ & $\bm{39.0\pm10.9}$ & $\bm{39.0\pm10.9}$ \\
TM Joint R &41.5 $\pm$ 10.0&41.5 $\pm$ 10.0& 41.5 $\pm$ 10.0& 41.5 $\pm$ 10.0&  $\bm{38.3\pm9.5}$&  $\bm{38.3\pm9.5}$&  $\bm{38.3\pm9.5}$& $\bm{38.3\pm9.5}$\\
Spinal Cord & 27.9 $\pm$ 8.3$^\dagger$& 27.7 $\pm$ 8.2&27.8 $\pm$ 8.3& 7.8 $\pm$ 2.9&  $\bm{34.7\pm6.9}$$^\dagger$ &  $\bm{34.7\pm6.9}$& $\bm{34.7\pm6.9}$& 11.7 $\pm$ 3.9\\
Mandible L & $\bm{78.0\pm4.9}$$^\dagger$& $\bm{78.0\pm5.2}$&68.0 $\pm$ 6.2& 47.0 $\pm$ 5.3& $\bm{66.7\pm6.1}$$^\dagger$ &  $\bm{66.7\pm5.2}$ &  46.7 $\pm$ 5.8&  22.9 $\pm$ 4.1\\
Mandible R & $\bm{71.4\pm4.0}$$^\dagger$ & 69.0 $\pm$ 4.4&58.5 $\pm$ 3.9& 40.2 $\pm$ 4.6& $\bm{66.0\pm4.7}$$^\dagger$ &  64.7 $\pm$ 4.9&   48.2 $\pm$ 5.0&  25.9 $\pm$ 3.2\\
\hline
Average & $\bm{39.1\pm9.7}$$^\dagger$ & 37.4 $\pm$ 9.7 & 35.8 $\pm$ 9.2 & 33.1 $\pm$ 9.1 & $\bm{47.3\pm7.4}$$^\dagger$ & 46.0 $\pm$ 7.4 & 43.0 $\pm$ 7.5 & 39.8 $\pm$ 7.2 \\ 
\bottomrule[1.5pt]
\end{tabular}}
\label{tab:8}
\end{table*}

%% file: tabel/tab9.tex
\begin{table*}[t!]
\centering
\caption{\textcolor{black}{Comparison of DSC scores $\uparrow$ (mean  $\pm$  std\%) between MedLAM-generated bounding box prompts and manually annotated prompts on the Totalsegmentator dataset.}}
\centering
\textcolor{black}{
\scalebox{0.95}{
\begin{tabular}{cccccc}
\toprule[1.5pt]
\multirow{2}*{Organs} & MedLAM & Manual Prompt & \multirow{2}*{Organs} & MedLAM & Manual Prompt \\ \cmidrule(lr){2-3} \cmidrule(lr){5-6}
~ & \multicolumn{2}{c}{SAM} & ~& \multicolumn{2}{c}{SAM}\\ \hline
Spleen & 71.2 $\pm$ 15.2 & $\bm{83.8 \pm 12.3}$ & Iliac Vena Left & 43.9 $\pm$ 11.5 & $\bm{68.7 \pm 5.0}$ \\
Kidney Right & 82.5 $\pm$ 7.0 & $\bm{90.3 \pm 2.6}$ & Iliac Vena Right & 58.7 $\pm$ 11.4 & $\bm{68.5 \pm 4.7}$ \\
Kidney Left & 80.6 $\pm$ 9.1 & $\bm{89.4 \pm 2.3}$ & Small Bowel & 38.0 $\pm$ 13.9 & $\bm{60.2 \pm 10.9}$ \\
Gallbladder & 56.5 $\pm$ 24.0 & $\bm{74.1 \pm 21.2}$ & Duodenum & 33.4 $\pm$ 21.0 & $\bm{53.9 \pm 11.7}$ \\
Liver & 70.5 $\pm$ 13.5 & $\bm{82.4 \pm 7.9}$ & Colon & 17.9 $\pm$ 8.4 & $\bm{37.1 \pm 4.9}$ \\
Stomach & 59.7 $\pm$ 15.3 & $\bm{77.6 \pm 11.7}$ & Rib Left 1 & 64.3 $\pm$ 12.2 & $\bm{75.9 \pm 6.4}$ \\
Aorta & 64.5 $\pm$ 17.6 & $\bm{75.5 \pm 11.2}$ & Rib Left 2 & 68.6 $\pm$ 11.5 & $\bm{73.1 \pm 9.1}$ \\
Inferior Vena Cava & 38.1 $\pm$ 13.6 & $\bm{77.4 \pm 7.7}$ & Rib Left 3 & 57.8 $\pm$ 12.8 & $\bm{71.9 \pm 8.6}$ \\
Portal Vein and Splenic vein & 19.5 $\pm$ 17.9 & $\bm{32.8 \pm 14.4}$ & Rib Left 4 & 56.1 $\pm$ 24.5 & $\bm{64.3 \pm 18.8}$ \\
Pancreas & 39.6 $\pm$ 18.8 & $\bm{60.9 \pm 13.5}$ & Rib Left 5 & 52.0 $\pm$ 27.3 & $\bm{65.9 \pm 20.0}$ \\
Adrenal Gland Right & 29.8 $\pm$ 15.4 & $\bm{49.0 \pm 11.6}$ & Rib Left 6 & 57.5 $\pm$ 15.6 & $\bm{74.6 \pm 8.0}$ \\
Adrenal Gland Left & 46.1 $\pm$ 13.1 & $\bm{56.0 \pm 10.6}$ & Rib Left 7 & 49.9 $\pm$ 21.6 & $\bm{73.0 \pm 14.1}$ \\
Lung Upper Lobe Left & 61.2 $\pm$ 12.7 & $\bm{85.5 \pm 10.1}$ & Rib Left 8 & 59.7 $\pm$ 16.8 & $\bm{72.5 \pm 7.5}$ \\
Lung Lower Lobe Left & 76.6 $\pm$ 19.2 & $\bm{85.0 \pm 13.1}$ & Rib Left 9 & 60.3 $\pm$ 18.9 & $\bm{72.8 \pm 12.7}$ \\
Lung Upper Lobe Right & 64.9 $\pm$ 20.8 & $\bm{84.3 \pm 18.7}$ & Rib Left 10 & 61.1 $\pm$ 13.8 & $\bm{73.9 \pm 4.7}$ \\
Lung Middle Lobe Right & 67.8 $\pm$ 16.1 & $\bm{79.0 \pm 12.1}$ & Rib Left 11 & 54.6 $\pm$ 9.9 & $\bm{72.4 \pm 4.6}$ \\
Lung Lower Lobe Right & 72.0 $\pm$ 27.3 & $\bm{81.1 \pm 20.2}$ & Rib Left 12 & 46.5 $\pm$ 14.1 & $\bm{66.5 \pm 6.8}$ \\
Vertebrae L5 & 72.2 $\pm$ 16.2 & $\bm{78.1 \pm 6.5}$ & Rib Right 1 & 53.6 $\pm$ 16.0 & $\bm{75.1 \pm 9.1}$ \\
Vertebrae L4 & 70.6 $\pm$ 14.1 & $\bm{78.9 \pm 7.7}$ & Rib Right 2 & 65.4 $\pm$ 14.3 & $\bm{76.1 \pm 7.3}$ \\
Vertebrae L3 & 70.1 $\pm$ 8.7 & $\bm{81.6 \pm 3.8}$ & Rib Right 3 & 48.2 $\pm$ 26.1 & $\bm{66.4 \pm 17.1}$ \\
Vertebrae L2 & 68.1 $\pm$ 9.8 & $\bm{80.0 \pm 5.8}$ & Rib Right 4 & 53.5 $\pm$ 19.8 & $\bm{68.2 \pm 17.0}$ \\
Vertebrae L1 & 71.2 $\pm$ 12.2 & $\bm{82.5 \pm 3.7}$ & Rib Right 5 & 51.3 $\pm$ 25.0 & $\bm{67.6 \pm 20.8}$ \\
Vertebrae T12 & 59.8 $\pm$ 16.2 & $\bm{78.2 \pm 13.2}$ & Rib Right 6 & 54.2 $\pm$ 16.3 & $\bm{76.7 \pm 7.1}$ \\
Vertebrae T11 & 68.1 $\pm$ 17.5 & $\bm{76.0 \pm 9.9}$ & Rib Right 7 & 57.2 $\pm$ 9.6 & $\bm{76.8 \pm 5.5}$ \\
Vertebrae T10 & 64.9 $\pm$ 7.4 & $\bm{77.2 \pm 4.8}$ & Rib Right 8 & 60.2 $\pm$ 22.4 & $\bm{72.5 \pm 12.9}$ \\
Vertebrae T9 & 69.2 $\pm$ 12.8 & $\bm{75.8 \pm 6.6}$ & Rib Right 9 & 51.5 $\pm$ 13.0 & $\bm{73.8 \pm 6.3}$ \\
Vertebrae T8 & 56.5 $\pm$ 19.0 & $\bm{67.9 \pm 16.8}$ & Rib Right 10 & 55.5 $\pm$ 8.9 & $\bm{74.7 \pm 4.6}$ \\
Vertebrae T7 & 57.3 $\pm$ 10.2 & $\bm{71.7 \pm 5.9}$ & Rib Right 11 & 50.7 $\pm$ 7.8 & $\bm{73.5 \pm 4.8}$ \\
Vertebrae T6 & 51.6 $\pm$ 20.3 & $\bm{61.9 \pm 16.1}$ & Rib Right 12 & 51.4 $\pm$ 14.4 & $\bm{66.6 \pm 6.8}$ \\
Vertebrae T5 & 48.2 $\pm$ 13.0 & $\bm{63.5 \pm 8.4}$ & Humerus Left & 70.4 $\pm$ 30.0 & $\bm{82.3 \pm 21.7}$ \\
Vertebrae T4 & 43.5 $\pm$ 24.5 & $\bm{56.0 \pm 17.3}$ & Humerus Right & 62.5 $\pm$ 34.8 & $\bm{75.4 \pm 28.3}$ \\
Vertebrae T3 & 58.7 $\pm$ 12.5 & $\bm{69.2 \pm 4.1}$ & Scapula Left & 67.7 $\pm$ 12.3 & $\bm{79.0 \pm 8.2}$ \\
Vertebrae T2 & 62.9 $\pm$ 14.7 & $\bm{71.2 \pm 12.3}$ & Scapula Right & 69.3 $\pm$ 18.0 & $\bm{80.4 \pm 9.6}$ \\
Vertebrae T1 & 65.0 $\pm$ 8.8 & $\bm{78.0 \pm 5.5}$ & Clavicula Left & 78.4 $\pm$ 11.2 & $\bm{89.1 \pm 3.3}$ \\
Vertebrae C7 & 62.9 $\pm$ 16.8 & $\bm{73.6 \pm 13.2}$ & Clavicula Right & 77.6 $\pm$ 15.3 & $\bm{86.5 \pm 8.2}$ \\
Vertebrae C6 & 67.3 $\pm$ 15.3 & $\bm{73.7 \pm 7.4}$ & Femur Left & 72.2 $\pm$ 14.3 & $\bm{90.4 \pm 4.8}$ \\
Vertebrae C5 & 62.3 $\pm$ 21.5 & $\bm{68.3 \pm 17.3}$ & Femur Right & 70.7 $\pm$ 19.5 & $\bm{87.1 \pm 16.1}$ \\
Vertebrae C4 & 56.4 $\pm$ 15.4 & $\bm{71.0 \pm 10.4}$ & Hip Left & 77.5 $\pm$ 9.9 & $\bm{88.9 \pm 3.6}$ \\
Vertebrae C3 & 64.5 $\pm$ 9.6 & $\bm{74.1 \pm 5.3}$ & Hip Right & 75.7 $\pm$ 8.4 & $\bm{89.7 \pm 3.6}$ \\
Vertebrae C2 & 68.6 $\pm$ 10.4 & $\bm{80.6 \pm 4.9}$ & Sacrum & 65.1 $\pm$ 12.9 & $\bm{82.8 \pm 5.5}$ \\
Vertebrae C1 & 58.9 $\pm$ 11.3 & $\bm{69.7 \pm 6.8}$ & Face & 52.9 $\pm$ 13.2 & $\bm{64.0 \pm 11.1}$ \\
Esophagus & 48.5 $\pm$ 8.1 & $\bm{68.7 \pm 5.8}$ & Gluteus Maximus Left & 57.1 $\pm$ 15.6 & $\bm{70.5 \pm 11.4}$ \\
Trachea & 77.5 $\pm$ 6.7 & $\bm{89.4 \pm 3.9}$ & Gluteus Maximus Right & 50.8 $\pm$ 20.7 & $\bm{68.8 \pm 16.5}$ \\
Heart Myocardium & 31.0 $\pm$ 14.7 & $\bm{52.6 \pm 10.0}$ & Gluteus Medius Left & 34.8 $\pm$ 18.3 & $\bm{51.2 \pm 14.7}$ \\
Heart Atrium Left & 63.6 $\pm$ 17.2 & $\bm{80.1 \pm 7.5}$ & Gluteus Medius Right & 31.9 $\pm$ 20.2 & $\bm{53.5 \pm 12.0}$ \\
Heart Ventricle Left & 57.9 $\pm$ 15.3 & $\bm{68.5 \pm 9.7}$ & Gluteus Minimus Left & 10.7 $\pm$ 17.9 & $\bm{23.8 \pm 9.7}$ \\
Heart Atrium Right & 58.0 $\pm$ 11.8 & $\bm{79.5 \pm 4.1}$ & Gluteus Minimus Right & 12.7 $\pm$ 18.3 & $\bm{21.7 \pm 12.1}$ \\
Heart Ventricle Right & 52.4 $\pm$ 10.5 & $\bm{74.9 \pm 4.9}$ & Autochthon Left & 53.5 $\pm$ 9.4 & $\bm{78.2 \pm 7.1}$ \\
Pulmonary Artery & 46.0 $\pm$ 24.2 & $\bm{54.8 \pm 17.9}$ & Autochthon Right & 54.3 $\pm$ 11.4 & $\bm{77.3 \pm 7.2}$ \\
Brain & 73.5 $\pm$ 12.8 & $\bm{81.1 \pm 8.5}$ & Iliopsoas Left & 43.0 $\pm$ 14.7 & $\bm{64.2 \pm 11.9}$ \\
Iliac Artery Left & 47.8 $\pm$ 13.1 & $\bm{67.4 \pm 9.4}$ & Iliopsoas Right & 58.2 $\pm$ 15.2 & $\bm{70.7 \pm 11.7}$ \\
Iliac Artery Right & 47.1 $\pm$ 13.0 & $\bm{66.0 \pm 10.8}$ & Urinary Bladder & 78.5 $\pm$ 16.1 & $\bm{84.9 \pm 8.0}$ \\
\bottomrule[1.5pt]
\end{tabular}}
}
\label{tab:9}
\end{table*}